\documentclass{article}

\usepackage[preprint]{neurips_2026}

\usepackage[utf8]{inputenc} % allow utf-8 input
\usepackage[T1]{fontenc}    % use 8-bit T1 fonts
\usepackage{enumitem}
\usepackage{url}            % simple URL typesetting
\usepackage{booktabs}       % professional-quality tables
\usepackage{makecell}       % for multi-line cells in tables
\usepackage{amsfonts}       % blackboard math symbols
\usepackage{nicefrac}       % compact symbols for 1/2, etc.
\usepackage{microtype}      % microtypography
\usepackage{xcolor}         % colors
\usepackage{algorithm}  
\usepackage{algorithmic} 
\usepackage[pdftex]{graphicx}
\usepackage{epstopdf}
\usepackage{amsmath}
\usepackage{amssymb}
\usepackage{amsthm}
\usepackage{multirow}
\usepackage{tabularx}
\usepackage{longtable}
\usepackage{caption}
\usepackage{float}
\usepackage{wrapfig}
\usepackage[skins, listings, breakable]{tcolorbox}
\usepackage{listings}
\usepackage{xcolor}
\usepackage{tikz}
\usepackage{authblk}
\usepackage{colortbl}

\usepackage[table]{xcolor}
\usepackage{pifont}
\usepackage{adjustbox}

\usepackage[utf8]{inputenc} % allow utf-8 input
\usepackage[T1]{fontenc}    % use 8-bit T1 fonts
\usepackage{amsfonts}       % blackboard math symbols
\usepackage{nicefrac}       % compact symbols for 1/2, etc.
\usepackage{microtype}      % microtypography
\usepackage{xcolor}         % colors
\usepackage[hidelinks]{hyperref} % load after the other packages

\usetikzlibrary{arrows.meta,positioning,calc,fit,backgrounds,shapes.geometric}

\usepackage{listings}

\definecolor{ORBenchHeader}{HTML}{F2F4F7}
\definecolor{ORBenchStripe}{HTML}{FAFBFD}
\definecolor{ORBenchGroup}{HTML}{EEF2F6}
\definecolor{ORBenchRule}{HTML}{3D4652}
\definecolor{ORBenchText}{HTML}{20252B}
\definecolor{ORBenchSubtle}{HTML}{6B7280}
\arrayrulecolor{ORBenchRule}
\newcolumntype{Y}{>{\raggedright\arraybackslash}X}
\newenvironment{orbenchtable}
  {\small\color{ORBenchText}\renewcommand{\arraystretch}{1.10}\setlength{\tabcolsep}{4.5pt}}
  {}
\newenvironment{orbenchcompact}
  {\scriptsize\color{ORBenchText}\renewcommand{\arraystretch}{1.06}\setlength{\tabcolsep}{3.2pt}}
  {}
\newcolumntype{D}{>{\raggedleft\arraybackslash}p{3.55em}}
\newcolumntype{E}{>{\centering\arraybackslash}p{3.55em}}

\newcommand{\orstripe}{\rowcolor{ORBenchStripe}}
\newcommand{\orgroupband}[2]{\rowcolor{ORBenchGroup}\multicolumn{#1}{@{}c@{}}{\rule{0pt}{1em}\textbf{#2}}}
\newcommand{\orbest}[1]{\textbf{#1}}
\newcommand{\orsecond}[1]{\underline{#1}}

\newcommand{\orcmark}{\textcolor{green!45!black}{\checkmark}}
\newcommand{\orxmark}{\textcolor{red!60!black}{\(\times\)}}
\newcommand{\orpart}{\textcolor{orange!75!black}{\(\triangle\)}}
\newcommand{\orcolhead}[1]{\textbf{#1}}

\newcommand{\orlogo}[1]{\raisebox{-0.14em}{\includegraphics[height=1.05em]{assets/logos/#1.png}}}
\newcommand{\oragent}[2]{\orlogo{#1}\hspace{0.35em}#2}
\newcommand{\orvectorlogo}[1]{\raisebox{-0.14em}{\includegraphics[height=1.05em]{assets/logos/#1.pdf}}}
\newcommand{\oropenaiagent}[1]{\orvectorlogo{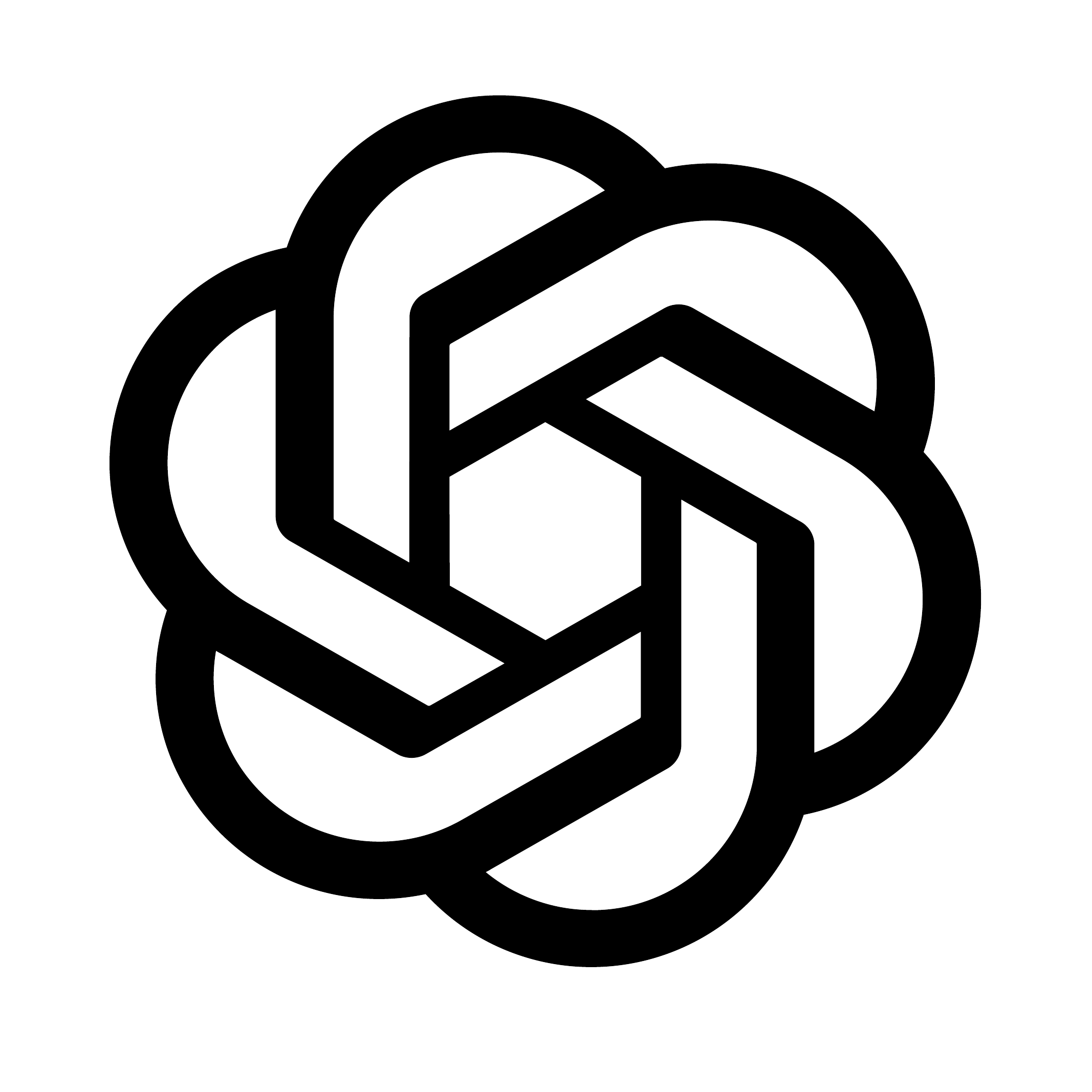}\hspace{0.35em}#1}
\newcommand{\orclaudemodelagent}[1]{\raisebox{-0.14em}{\includegraphics[height=0.95em]{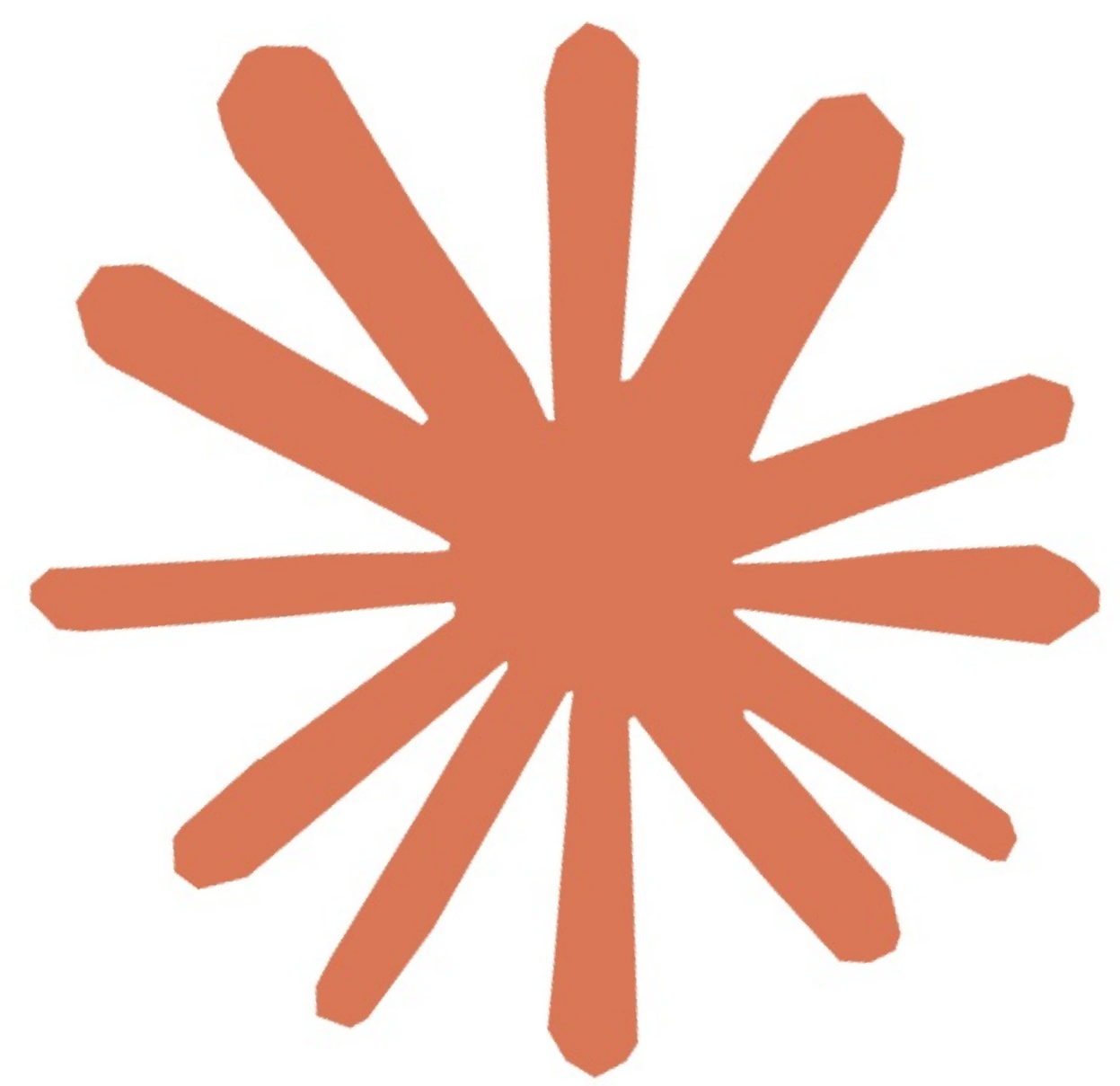}}\hspace{0.35em}#1}
\newcommand{\orzhipuagent}[1]{\orvectorlogo{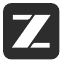}\hspace{0.35em}#1}

\definecolor{ORCaseHeader}{HTML}{D9D9E8}
\definecolor{ORCaseBorder}{HTML}{B7B9CE}
\definecolor{ORCaseFail}{HTML}{A40000}
\definecolor{ORCaseFailBack}{HTML}{FFF2F2}
\definecolor{ORCaseSuccess}{HTML}{0B7A16}
\definecolor{ORCaseSuccessBack}{HTML}{F1FFF1}
\definecolor{ORCaseInfo}{HTML}{315F86}
\definecolor{ORCaseInfoBack}{HTML}{F3F7FB}
\definecolor{ORCaseAccent}{HTML}{DCE8F2}
\lstdefinestyle{orbenchcode}{
  basicstyle=\ttfamily\scriptsize,
  keywordstyle=\color{blue!70!black}\bfseries,
  commentstyle=\color{black!45}\itshape,
  stringstyle=\color{green!40!black},
  showstringspaces=false,
  breaklines=true,
  columns=fullflexible,
  keepspaces=true,
  frame=none,
  xleftmargin=1pt,
  aboveskip=2pt,
  belowskip=2pt
}
\newtcolorbox{orappendixbox}[1]{
  enhanced,
  breakable,
  colback=white,
  colframe=ORCaseBorder,
  colbacktitle=ORCaseHeader,
  coltitle=black,
  fonttitle=\bfseries,
  title={#1},
  boxrule=0.45pt,
  arc=0pt,
  left=6pt,
  right=6pt,
  top=5pt,
  bottom=5pt,
  before skip=5pt,
  after skip=5pt
}
\newtcolorbox{orfailurebox}[1]{
  enhanced,
  colback=ORCaseFailBack,
  colframe=ORCaseFail,
  colbacktitle=red!10,
  coltitle=ORCaseFail,
  fonttitle=\bfseries\small,
  fontupper=\small,
  before upper={\raggedright},
  title={#1},
  boxrule=0.7pt,
  arc=0pt,
  left=6pt,
  right=6pt,
  top=4pt,
  bottom=4pt
}
\newtcolorbox{orsuccessbox}[1]{
  enhanced,
  colback=ORCaseSuccessBack,
  colframe=ORCaseSuccess,
  colbacktitle=green!10,
  coltitle=ORCaseSuccess,
  fonttitle=\bfseries\small,
  fontupper=\small,
  before upper={\raggedright},
  title={#1},
  boxrule=0.7pt,
  arc=0pt,
  left=6pt,
  right=6pt,
  top=4pt,
  bottom=4pt
}
\newtcolorbox{oroverviewbox}[1]{
  enhanced,
  colback=ORCaseInfoBack,
  colframe=ORCaseInfo,
  colbacktitle=ORCaseAccent,
  coltitle=black,
  fonttitle=\bfseries\small,
  fontupper=\small,
  before upper={\raggedright},
  title={#1},
  boxrule=0.6pt,
  arc=1pt,
  left=7pt,
  right=7pt,
  top=6pt,
  bottom=6pt,
  before skip=6pt,
  after skip=7pt
}
\newtcolorbox{ortrajectorybox}[1]{
  enhanced,
  colback=white,
  colframe=ORCaseInfo!75,
  colbacktitle=ORCaseInfoBack,
  coltitle=black,
  fonttitle=\bfseries\small,
  fontupper=\small,
  before upper={\raggedright},
  title={#1},
  boxrule=0.55pt,
  arc=0pt,
  left=6pt,
  right=6pt,
  top=5pt,
  bottom=5pt,
  before skip=5pt,
  after skip=5pt
}
\newtcolorbox{ororaclebox}{
  enhanced,
  colback=white,
  colframe=ORCaseInfo!75,
  colbacktitle=ORCaseInfoBack,
  coltitle=black,
  fonttitle=\bfseries\footnotesize,
  fontupper=\footnotesize,
  before upper={\raggedright\setlength{\medskipamount}{3pt plus 1pt minus 1pt}},
  title={Selected major hard constraints checked by the verifier},
  boxrule=0.55pt,
  arc=1pt,
  left=5pt,
  right=5pt,
  top=4pt,
  bottom=4pt,
  before skip=4pt,
  after skip=5pt
}
\newcommand{\orcheckcell}[3]{%
  \begin{minipage}[t]{0.485\linewidth}
  \vspace{0pt}\raggedright
  \tikz[baseline=(badge.base)]{
    \node[circle,fill=ORCaseInfo,text=white,inner sep=0pt,
      minimum size=1.35em,font=\bfseries\tiny] (badge) {#1};
  }%
  \hspace{1pt}\textbf{#2}\par
  {\scriptsize\setlength{\baselineskip}{7.4pt}#3}
  \end{minipage}%
}
\newcommand{\ordimbar}[1]{%
  \begin{tikzpicture}[baseline=-0.45ex,x=0.82em,y=0.72em]
    \foreach \i in {1,2,3}{
      \ifnum\i>#1
        \fill[ORBenchHeader] (\i-1,0) rectangle ++(0.78,0.72);
        \draw[ORCaseBorder,line width=0.25pt] (\i-1,0) rectangle ++(0.78,0.72);
      \else
        \fill[ORCaseInfo] (\i-1,0) rectangle ++(0.78,0.72);
      \fi
    }
  \end{tikzpicture}%
}

\PassOptionsToPackage{number,compress}{natbib}

\definecolor{ORBenchBlue}{RGB}{20,76,150}

\title{ORAgentBench: Can LLM Agents Solve Challenging Operations Research Tasks End to End?}

\author{
\bf  Jiajun Li$^{1}$ \quad Mingshu Cai$^{2}$ \quad Yixuan Li$^{3}$ \quad Yu Ding$^{1,4}$ \\
\bf 
Ran Hou$^{1}$ \quad Guanyu Nie$^{4}$ \quad Xiongwei Han$^{4}$ \quad Wanyuan Wang$^{1}$\thanks{Corresponding author. E-mail: wywang@seu.edu.cn}
\\
\vspace{0.5em}
$^{1}$School of Computer Science and Engineering, Southeast University \quad \\
$^{2}$Waseda University \quad
$^{3}$Nanyang Technological University \quad \\
$^{4}$Huawei Noah's Ark Lab
}

\begin{document}

\maketitle

\begin{abstract}
Large language models are increasingly deployed as autonomous agents for multi-step tasks in executable environments, yet their ability to perform realistic operations research (OR) work remains unclear. Existing OR evaluations often decouple modeling from solving, rely on pre-formalized or text-only instances, and rarely test the full workflow from operational artifacts to validated decisions.
In this work, we introduce \textbf{ORAgentBench}, an execution-grounded benchmark for evaluating autonomous agents on challenging end-to-end operations research tasks.
It contains 107 human-reviewed tasks across diverse operational scenarios, each packaged in an isolated environment with a natural-language brief, multi-file data, configuration artifacts, and a required submission schema. 
Agents must write and run solution code, and their submissions are evaluated by hidden validators for schema validity, hard-constraint feasibility, and normalized objective quality. Experiments with fourteen frontier agent-model configurations show that current agents remain far from reliable OR practice. The best agent passes only 35.51\% of all tasks and 20.59\% of hard tasks, and many feasible submissions still fall below the required quality threshold. Failure analysis further shows that errors are dominated by strategic weaknesses, including missed operational rules, brittle formulations, weak feasible-solution construction, and insufficient solution improvement. OR-specific procedural skills increase hard-task feasibility, but do not reliably improve solution quality or pass rate.
These results suggest that progress in OR agents requires moving beyond plausible optimization code toward dependable, high-quality operational decision-making.
Code is available at \href{https://github.com/ORAgentBench/ORAgentBench}{\textcolor{ORBenchBlue}{\textbf{ORAgentBench}}}.
\end{abstract}

% \begin{figure}[h]
%     \centering
%     \includegraphics[width=\linewidth]{assets/figures/abstract_benchmark_summary.pdf}
%     \caption{Overall ORAgentBench outcomes by model-agent harness.}
%     \label{fig:abstract-benchmark-summary}
% \end{figure}

\section{Introduction}
\label{intro}

Operations research (OR) underpins high-value operational decisions in freight routing \citep{toth2014vehicle}, workforce scheduling \citep{pinedo2016scheduling}, production planning \citep{pochet2006production}, energy dispatch \citep{conejo2010decision}, inventory control \citep{zipkin2000foundations}, and public-service allocation \citep{daskin2013network}. 
Even small improvements in feasibility, utilization, delay, cost, or service reliability can yield substantial economic and social impact. Yet realistic OR work is rarely solved by formulation alone. 
Practical mathematical programming is an end-to-end engineering workflow that turns ambiguous operational needs into executable or auditable decision artifacts through iterative data reconciliation, model design, implementation, validation, and revision against operational constraints \citep{williams2013modelbuilding}. 
This workflow typically requires coordinated expertise from domain specialists, data experts, modeling and solving experts, and software engineers, making real OR projects costly, time-consuming, and failure-prone.

\begin{figure}[htbp]
    \vspace{-0.8em}
    \centering
    \includegraphics[width=\linewidth]{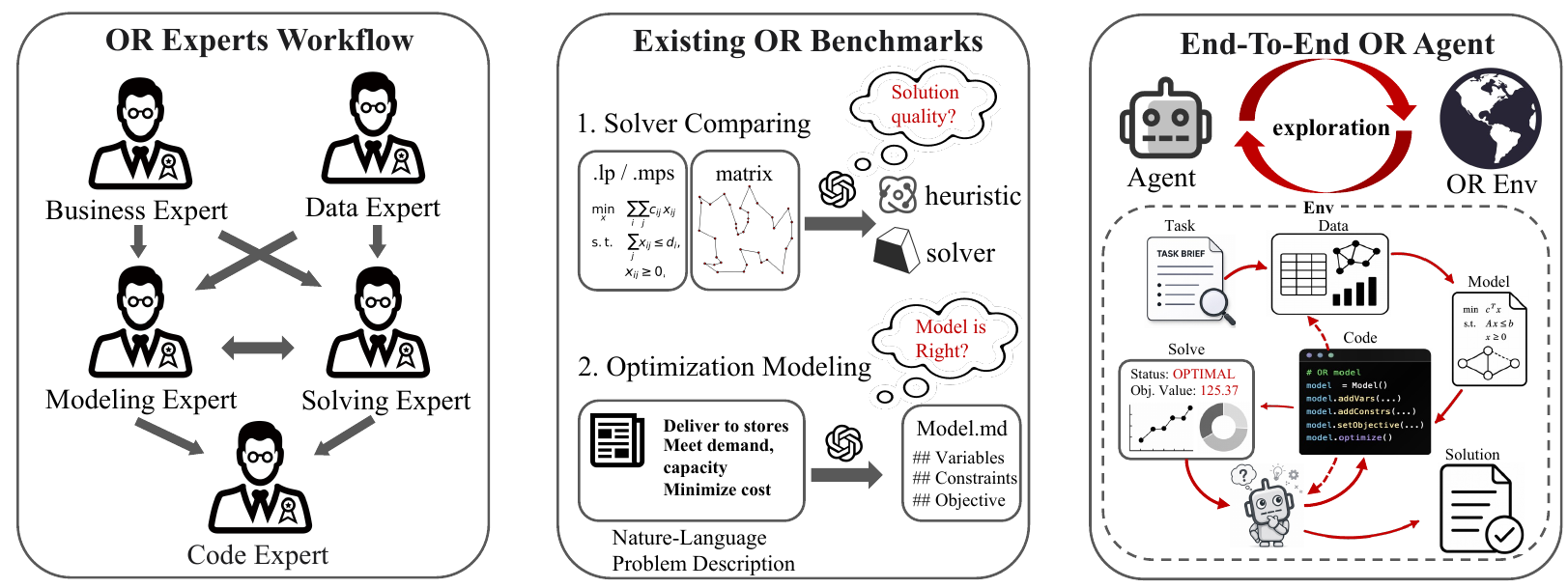}
    \caption{Motivation of ORAgentBench. Realistic OR work relies on coordinated multi-expert workflows; existing benchmarks test isolated stages, while ORAgentBench evaluates the complete executable OR workflow.
    }
    \vspace{-0.8em}
    \label{fig:intro}
\end{figure}

Recent advances in foundation models have substantially expanded the operational scope of AI agents. 
Agents have shown early progress across execution-grounded domains, including software engineering \citep{jimenez2024swebench,yang2026programbench,zhang2026repozero}, web and computer use \begin{NoHyper}\citep{zhou2024webarena,xie2024osworld}\end{NoHyper}, automated scientific and research workflows \citep{gottweis2026coscientist,schmidgall2025agentlaboratory}, and broader real-world agent settings \citep{tang2026workspacebench,ye2026claweval}.
Yet their capabilities in operations research remain unclear, especially for realistic end-to-end OR tasks. 
As illustrated in Figure~\ref{fig:intro}, this uncertainty reflects a mismatch between existing OR-agent evaluations and realistic OR workflows. Current OR benchmarks capture important parts of this workflow, but not the full end-to-end process. 
Solver-centric benchmarks evaluate algorithms on already-formalized optimization instances \citep{gleixner2021miplib,stuckey2014minizincchallenge}, leaving modeling from operational artifacts outside the evaluation. 
Modeling-oriented benchmarks test whether models can generate formulations, solver code, or numerical answers from problem descriptions \citep{ramamonjison2023nl4opt,huang2024orlm,ahmaditeshnizi2024optimus}, but typically do not evaluate whether agents can reconcile data, execute and repair code, and produce validated decisions under runtime constraints. 
Recent agentic OR systems highlight the promise of workflow-level optimization agents \citep{zhang2025orllmagent,song2026nemoexecutionawareoptimizationmodeling}, but still do not establish whether agents can complete realistic OR work from operational artifacts to final decision artifacts.
These gaps expose a deeper issue: existing evaluations often separate modeling from solving, whereas realistic OR requires joint design. 
An OR agent must choose a \textbf{modeling strategy} $\phi_{\theta,\tau}$ that faithfully captures requirements without producing oversized or brittle formulations, together with a \textbf{solving strategy} $\psi_{\theta,\tau}$ that obtains good feasible solutions under time limits through exact, heuristic, repair-based, or hybrid methods. 
Benchmarks limited to formulations or pre-formalized instances therefore cannot distinguish agents that write plausible equations from those that complete operational OR work.

% \begin{figure}[t]
%     \centering
%     \includegraphics[width=\linewidth]{assets/figures/task-construct.pdf}
%     \caption{The task construction pipeline of \textbf{ORAgentBench}.}
%     \label{fig:or-agent-gap}
% \end{figure}

To address this gap, we introduce \textsc{ORAgentBench}, a benchmark of 107 executable tasks for evaluating agents on realistic OR workflows. 
Tasks are constructed from OR papers, real-world industrial scenarios, and benchmark seed instances, and packaged as isolated environments. 
Agents are evaluated by their ability to produce feasible, high-quality decision artifacts under these constraints.
In summary, our contributions are:
\begin{itemize}[leftmargin=*]
    \item \textbf{\textsc{ORAgentBench}.} We introduce the first benchmark for realistic end-to-end OR-agent evaluation, covering 107 executable tasks.
    
    \item \textbf{Comprehensive benchmark evaluation.} We evaluate frontier model--harness combinations and our results show that current agents still struggle to reliably complete realistic OR tasks end to end.
    
    \item \textbf{Analysis of skill-enhanced OR agents.} We study how domain-specific skill guidance affects agent performance. The results show that skills reshape the failure profile rather than uniformly improving performance: they improve hard-task feasibility but not reliably quality or pass rate.
\end{itemize}

\begin{table*}[t]
    \centering
    \caption{Comparison with existing OR and Agent benchmarks on multiple dimensions: 
    \textbf{Multi-files} denotes multiple input artifacts; 
    \textbf{Multi-turns} denotes simulated user interaction; 
    \textbf{Long-horizon} denotes extended task trajectories; 
    \textbf{Model} denotes optimization modeling; 
    \textbf{Execution} denotes code or solver execution; 
    \textbf{Quality Eval.} denotes solution-quality evaluation; 
    and \textbf{Isolated env.} denotes sandboxed execution environment. 
    \orcmark{} = full support; \orpart{} = partial support; \orxmark{} = absent.}
    \label{tab:benchmark-comparison}
    \vspace{-0.5em}

    \begin{adjustbox}{max width=\linewidth}
    \renewcommand{\arraystretch}{1.12}
    \setlength{\tabcolsep}{7pt}
    \begin{tabular}{@{}l@{\extracolsep{\fill}}ccccccc@{}}
        \toprule
        \textbf{Benchmark} 
        & \textbf{Multi-files} 
        & \textbf{Multi-turns} 
        & \textbf{Long-horizon} 
        & \textbf{Model} 
        & \textbf{Execution} 
        & \textbf{Quality Eval.} 
        & \textbf{Isolated env.} \\
        \midrule

        NL4Opt~\citep{ramamonjison2023nl4opt} 
        & \orxmark & \orxmark & \orxmark & \orcmark & \orcmark & \orxmark & \orxmark \\

        IndustryOR~\citep{huang2024orlm} 
        & \orxmark & \orxmark & \orxmark & \orcmark & \orcmark & \orxmark & \orxmark \\

        MIPLIB-NL~\citep{li2026miplibnl} 
        & \orcmark & \orxmark & \orxmark & \orcmark & \orcmark & \orxmark & \orxmark \\

        MIPLIB~\citep{gleixner2021miplib} 
        & \orpart & \orxmark & \orxmark & \orxmark & \orcmark & \orcmark & \orxmark \\

        CVRPLIB~\citep{uchoa2017cvrplib} 
        & \orpart & \orxmark & \orxmark & \orxmark & \orcmark & \orcmark & \orxmark \\

        CO-Bench~\citep{sun2025cobench} 
        & \orpart & \orxmark & \orcmark & \orxmark & \orcmark & \orcmark & \orpart \\

        MLE-Bench~\citep{chan2024mlebench} 
        & \orcmark & \orxmark & \orcmark & \orxmark & \orcmark & \orcmark & \orcmark \\

        Terminal-Bench~\citep{merrill2026terminalbench} 
        & \orcmark & \orxmark & \orcmark & \orxmark & \orcmark & \orpart & \orcmark \\

        SWE-Bench~\citep{jimenez2024swebench} 
        & \orcmark & \orxmark & \orcmark & \orxmark & \orcmark & \orxmark & \orcmark \\

        \midrule
        % \rowcolor{gray!18}
        \textbf{ORAgentBench (Ours)}
        & \orcmark & \orcmark & \orcmark & \orcmark & \orcmark & \orcmark & \orcmark \\
        \bottomrule
    \end{tabular}
    \end{adjustbox}

    \vspace{-1.0em}
\end{table*}

\section{Related Work}
\label{related}

\subsection{LLMs for Optimization Modeling}

LLMs for optimization modeling have progressed from equation generation toward executable modeling workflows. 
Prior work improves formulation and solver-code generation through domain training and synthetic data \citep{huang2024orlm,lu2025optmathscalablebidirectionaldata}, and extends modeling into iterative execution, repair, debugging, and coding-agent workflows \citep{ahmaditeshnizi2024optimus,zhang2025orllmagent,song2026nemoexecutionawareoptimizationmodeling}. 
However, these works mainly test whether models can produce plausible formulations or solver code, rather than whether agents can choose abstractions and formulations that are strong and effective. 
\textsc{ORAgentBench} targets this missing capability by evaluating modeling choices through executable solutions and validated decision quality in realistic OR workflows.

\subsection{OR and Agent Benchmarks}
Existing benchmarks evaluate important parts of OR-agent capability, but not the full artifact-to-decision workflow required by realistic OR work. 
Classical OR benchmarks, such as MIPLIB and CVRPLIB \citep{gleixner2021miplib,uchoa2017cvrplib}, rigorously evaluate solvers or algorithms on already-formalized instances, leaving modeling from operational artifacts outside the evaluation.
LLM-oriented OR benchmarks move evaluation closer to language-based modeling and solver-assisted decision generation. 
NL4Opt and Mamo evaluate natural-language mathematical modeling with solver-backed checks \citep{ramamonjison2023nl4opt,huang2024mamo}; 
IndustryOR, OptiBench, OptMATH, and MIPLIB-NL extend this line toward practical OR, solver-assisted end-to-end solving, scalable synthetic construction, and industrial-scale mixed-integer programs \citep{huang2024orlm,yang2025optibench,lu2025optmathscalablebidirectionaldata,li2026miplibnl}. 
For combinatorial optimization, CO-Bench evaluates agentic algorithm search \citep{sun2025cobench}, while NLCO studies direct natural-language reasoning over CO problems \citep{jiang2026nlco}. 
Together, these benchmarks cover key language-to-modeling, solver-assisted solving, and CO reasoning abilities, but not the full workflow from operational artifacts to validated decisions.
General agent benchmarks evaluate execution-grounded tasks in software engineering \citep{jimenez2024swebench,yang2025swebenchmultimodal}, machine-learning engineering \citep{chan2024mlebench,huang2024mlagentbench}, web interaction \citep{zhou2024webarena,koh2024visualwebarena}, desktop control \citep{xie2024osworld,bonatti2024windowsagentarena}, and broader heterogeneous settings \citep{liu2024agentbench,froger2026gaia2}.
However, their outputs are typically patches, actions, reports, or task completions, rather than constrained operational decisions evaluated by feasibility and objective quality.
\textsc{ORAgentBench} fills this gap by combining realistic multi-file execution environments with OR-style validation of constrained, objective-driven decision artifacts, as summarized in Table~\ref{tab:benchmark-comparison}.

% Existing OR benchmarks offer useful but limited coverage of realistic OR work. 
% Solver-centric benchmarks such as MIPLIB~\citep{gleixner2021miplib} and the MiniZinc Challenge~\citep{stuckey2014minizincchallenge} measure performance on already-formalized optimization instances, leaving modeling-from-artifacts outside the evaluation. 
% Language-based modeling benchmarks such as NL4Opt~\citep{ramamonjison2023nl4opt}, ORLM/IndustryOR~\citep{tang2024orlm}, and OptiMUS~\citep{ahmaditeshnizi2024optimus} evaluate formulation or solver-code generation, but not the downstream process of producing validated decisions under data and runtime constraints. 
% Beyond benchmarks, agentic OR systems such as OR-LLM-Agent~\citep{orllmagent2025} and NEMO~\citep{song2026nemoexecutionawareoptimizationmodeling} decompose optimization workflows into modeling, code generation, execution, validation, and repair, while CO-Bench~\citep{sun2025cobench} evaluates agentic algorithm search for combinatorial optimization. 
% Together, these works show the promise of agentic optimization, but existing evaluations still do not fully test whether an agent can inspect heterogeneous operational artifacts, reconcile data, design both modeling and solving strategies, execute and repair code, and submit a decision artifact to a hidden validator.

% \input{sections/preliminaries}

\section{ORAgentBench}
\label{method:bench}

\subsection{Problem Definition}
\label{sec:task-definition}

To bridge the gap between simplified text-to-formulation evaluation and the operational rigor of real-world deployment, we formally define an \textbf{end-to-end operations research (OR) task} as:
\begin{equation}
    \tau = (\mathcal{D}, \mathcal{C}, q, \mathcal{E}, \mathcal{V}),
\end{equation}
where $\mathcal{D}$ and $\mathcal{C}$ denote the input data and configuration, $q$ is a natural-language operational brief, $\mathcal{E}$ represents an isolated sandbox execution environment, and $\mathcal{V}$ is a hidden task-specific validator. 

Given the public task packet $I_{\tau}=(\mathcal{D},\mathcal{C},q)$, an autonomous OR agent $A_{\theta}$ parameterised by $\theta$ generates an executable program:
\begin{equation}
    p_{\theta}=A_{\theta}(I_{\tau}).
\end{equation}
Running $p_{\theta}$ within the environment $\mathcal{E}$ yields a concrete decision artifact:
\begin{equation}
    s = \mathrm{Exec}_{\mathcal{E}}(p_{\theta};\mathcal{D},\mathcal{C}),
\end{equation}
which is subsequently quantified by the validator $\mathcal{V}$ via an objective score:
\begin{equation}
    \mathrm{Score}(\theta,\tau)=\mathcal{V}(s).
\end{equation}
The validator $\mathcal{V}$ rigorously evaluates $s$ across three progressive dimensions: output-schema validity, hard constraint feasibility, and objective quality. Concretely, we view the program $p_{\theta}$ as an observable realization driven by two coupled, internal decisions: a \textbf{latent modeling strategy} ($\phi_{\theta,\tau} \in \Phi$) and a \textbf{latent solving strategy} ($\psi_{\theta,\tau} \in \Psi$). This dual-strategy generation process can be conceptually represented as:
\begin{equation}
    p_{\theta} = A_{\theta}(I_{\tau} \mid \phi_{\theta,\tau}, \psi_{\theta,\tau}),
\end{equation}
where $\phi_{\theta,\tau}$ and $\psi_{\theta,\tau}$ signify implicit cognitive choices rather than required external submissions. This formulation fundamentally distinguishes \texttt{ORAgentBench} from traditional natural-language-to-formulation benchmarks. Previous evaluation paradigms primarily test a model's translation capability. In contrast, realistic OR applications present a dual challenge: the agent must implicitly co-design an effective problem representation ($\phi_{\theta,\tau}$) and an efficient computational procedure ($\psi_{\theta,\tau}$), executing both seamlessly as programmatic code that delivers high-quality operational decisions.

\begin{figure}[t]
    \centering
    \includegraphics[width=\linewidth]{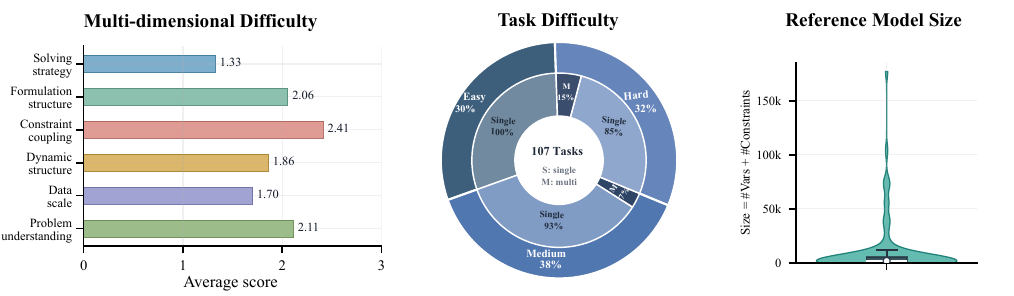}
    \caption{
    Benchmark analysis overview.
    Left: average pressure of the six construction-time difficulty dimensions.
    Middle: easy/medium/hard split, with each level decomposed into single-turn and  multi-turns tasks.
    Right: distribution of reference-model size, measured as variables plus constraints.
    }
    \vspace{-0.6em}
    \label{fig:benchmark-composition-overview}
\end{figure}

\subsection{Design Principles}
\label{sec:design-principles}
% ORAgentBench is built around three design principles.

\paragraph{P1: Difficulty should be controlled by modeling and solving burden.}
Task difficulty should reflect the reasoning required to construct an effective model and obtain a high-quality solution, rather than instance size alone.
Small instances may be difficult when they require subtle interpretation or tightly coupled reasoning, whereas larger instances may be easier when they admit standard, well-structured formulations.
We therefore control difficulty prospectively through six construction-time dimensions: solving strategy requirement, formulation structure, constraint coupling, dynamic structure, data scale, and problem understanding.
Together, these dimensions distinguish the source of difficulty and prevent scale from serving as a misleading proxy for task complexity.

\paragraph{P2: Scenarios should be operationally grounded and automatically verifiable.}
Each scenario must encode a concrete operational decision problem with an explicit objective, meaningful resource constraints, and substantive OR structure.
To ensure valid and reproducible evaluation, all information needed to solve the task must be contained in the provided materials; tasks must not rely on live external information or unstated domain knowledge.
The resulting decision artifact must be deterministically verifiable and substantively distinct from existing tasks.
We exclude scenarios whose primary challenge is prediction, simulation, or open-ended analysis rather than optimization.

\paragraph{P3: Data should reflect realistic business artifacts.}
Tasks expose structured, multi-file artifacts rather than a fully normalized model instance.
The data preserve the dependencies and operational semantics needed to formulate the problem, while numerical values and constraint tightness are calibrated to avoid trivial solutions.
All artifacts are fixed for reproducibility, requiring agents to infer a coherent optimization model from the provided evidence rather than apply a prompt-level template.

\paragraph{Benchmark analysis.}
Applying the above principles, ORAgentBench is designed to cover diverse end-to-end OR workflows rather than surface-level prompt variants. 
Figure~\ref{fig:benchmark-composition-overview} analyzes the benchmark composition from three perspectives: construction-time difficulty, task mode, and reference-model scale. 
The results show that ORAgentBench varies not only in instance size, but also in the modeling, solving, and artifact-understanding demands imposed on agents.

\subsection{Task Construction}
\label{sec:task-construction}

\begin{figure}[t]
    \centering
    \includegraphics[width=\linewidth]{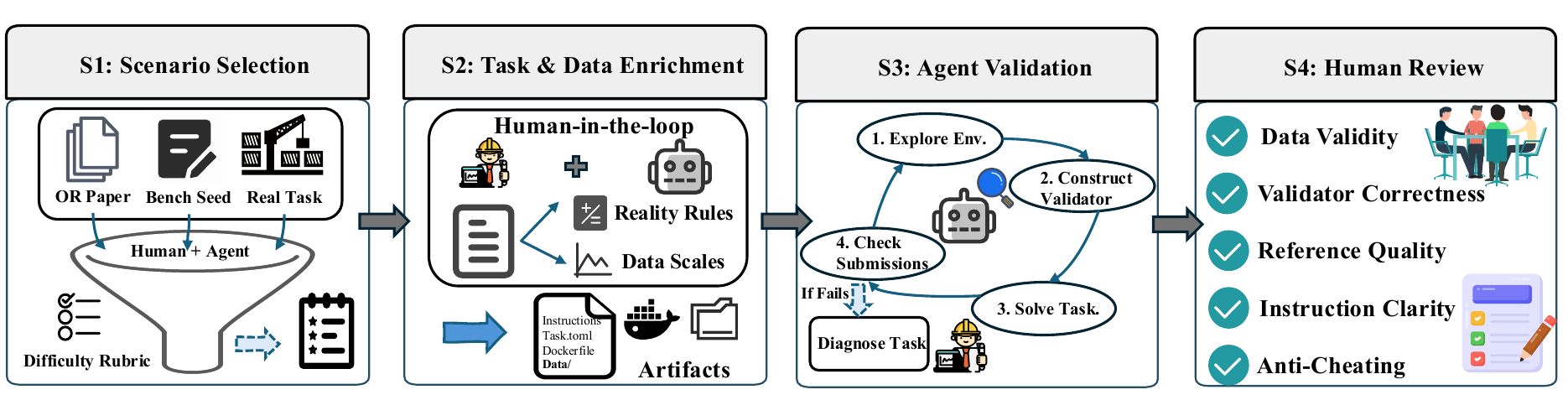}
    \caption{The task construction pipeline of \textbf{ORAgentBench}.}
    \label{fig:construction-pipeline}
\end{figure}

Figure~\ref{fig:construction-pipeline} summarizes our task construction pipeline.
Starting from OR papers, benchmark seeds, and real operational scenarios, the pipeline turns candidate settings into executable, reproducible, and automatically evaluable benchmark tasks through four stages: scenario selection, task and data enrichment, agent-assisted validation, and human review.

\paragraph{Stage 1: scenario selection and difficulty framing.}
We collect candidate scenarios from OR papers, existing benchmark seeds, public datasets, application studies, and industrial planning patterns.
Human designers and agents jointly screen these candidates by identifying the core decision process, objective, resource limits, data requirements, and dominant OR structure.
We retain scenarios that are concrete, reproducible, executable in an isolated environment, and amenable to deterministic validation.
We exclude cases that are mainly prediction, simulation, open-ended consulting, duplicate, dependent on live APIs, or difficult to score automatically.
For each retained scenario, we assign a target difficulty profile using our construction-time rubric, covering solving strategy, formulation structure, constraint coupling, dynamic structure, data scale, and problem understanding.

\paragraph{Stage 2: task and data enrichment.}
We expand each selected scenario into an executable benchmark instance through a human-in-the-loop process, packaging it with the instruction, task configuration, Docker environment, and structured data files. During this process, we enrich the task with realistic operational rules, interacting constraints, and calibrated data scales, so that difficulty arises from substantive modeling and solving decisions rather than prompt length or file size. 
For data construction, we synthesize and extend data from original sources whenever available, and supplement missing parts with operationally plausible synthetic data, including realistic identifiers, non-round values, heterogeneous cases, and cross-file dependencies. Public files provide all information needed for solving, while hidden files are reserved for validation and scoring.

\paragraph{Stage 3: agent-assisted validation.}
After the task package is drafted, we run an agent-assisted validation loop.
The agent explores the environment, constructs a deterministic validator, solves the task using only public information, and checks the submitted solution.
When validation fails, we diagnose whether the issue comes from the solution, instruction, data, validator, or task design.
We then revise the corresponding component and repeat the loop until the task has a valid reference solution and a reliable validator.

\paragraph{Stage 4: human review.}
Finally, human reviewers inspect the complete task package, including the instruction, data, environment, reference solution, solving method, and validator.
The review focuses on data validity, validator correctness, reference quality, instruction clarity, leakage risks, shortcut opportunities, and anti-cheating issues.
Reviewers also check whether the public requirements are consistent with hidden validation logic and whether the task remains realistic without exposing the underlying formulation too directly.
Only tasks that pass this review are included in the benchmark.

% \subsection{Benchmark Composition}
% \label{sec:benchmark-analysis}

% ORAgentBench currently contains 107 executable tasks, each with an agent-facing brief, public data files, a submission schema, a hidden deterministic validator, and a six-dimensional construction profile. The suite covers both standard OR families and realistic hybrids, including routing, scheduling, assignment, network/location, production/allocation, inventory/supply, energy, healthcare, emergency response, and resource planning. Beyond the aggregate views in Figure~\ref{fig:benchmark-composition-overview}, the scenario distribution spans industrial operations, transportation and mobility, healthcare and emergency response, energy, water, environment, digital infrastructure, and service operations. This breadth is important because the benchmark tests whether agents can adapt their modeling and solving strategies across operational artifacts rather than memorize a single formulation template.

\subsection{Scoring Protocol}
\label{sec:scoring-protocol}

Each task is evaluated by a hidden validator that maps a submitted decision artifact $s$ to two scores:
\begin{equation}
    \mathcal{V}(\tau,s)=\big(F(\tau,s), q(\tau,s)\big),
\end{equation}
where $F(\tau,s)\in[0,1]$ measures operational validity and $q(\tau,s)\in[0,1]$ measures objective quality.
Feasibility is a hard gate: violations of the required schema, domains, or hard operational constraints set $F(\tau,s)=0$ and yield zero task credit.

For feasible submissions, quality is normalized relative to a verified reference objective $R$ and, when available, a valid global bound $B$. Let $g_\tau(s)$ be the signed improvement of the submitted objective over $R$, and let $W_\tau$ be the reference-to-bound width (like 0.05\%), with signs adjusted for minimization or maximization. When $W_\tau$ is numerically meaningful, we compute
\begin{equation}
    q(\tau,s)=\frac{1}{2}\,
    \operatorname{clip}\!\left(
    1+\frac{g_\tau(s)}{W_\tau},\,0,\,2
    \right).
\end{equation}
If no meaningful bound is available, the validator uses a tolerance-scaled reference fallback. Appendix~\ref{app:scoring-details} gives the objective-direction conventions, fallback rule, and raw Harbor reward. We define the binary pass indicator as
\begin{equation}
    \mathrm{Pass}(\tau,s)=
    \mathbb{I}\!\left[
    F(\tau,s)>0
    \land
    q(\tau,s)>0.4
    \right].
\end{equation}
A task is passed only when the submission is feasible and reaches the quality threshold.

\section{Experiments}
\label{experiments}

Our experiments are designed to diagnose end-to-end OR-agent capability, rather than merely rank models by a single answer score. We organize the empirical analysis around four research questions. 
\textbf{RQ1}: How well do frontier model-agent stacks solve ORAgentBench? 
\textbf{RQ2}: How do cost and runtime vary with task difficulty and success? 
\textbf{RQ3}: Which failure modes dominate, and do they arise primarily from modeling-strategy or solving-strategy failures? 
\textbf{RQ4}: Can OR-specific skills improve end-to-end OR-agent performance?

\subsection{Experimental Setup}
\label{sec:eval-protocol}

\paragraph{Models.}
We evaluate fourteen model-agent rows on all 107 tasks.
The backends cover MiniMax-M2.7 \citep{minimax2026m2series}, Kimi K2.6 \citep{moonshot2026kimik26}, MiMo V2.5 Pro \citep{xiaomi2026mimov25}, Qwen3.5/3.6 Plus \citep{alibabacloud2026modelstudio}, DeepSeek V4 Flash/Pro \citep{deepseek2026v4preview}, GLM 5/5.1 \citep{glmteam2026glm5,zhipu2026glm51modelcard}, Claude Sonnet/Opus 4.6 \citep{anthropic2026claude46systemcards}, and three OpenAI models: GPT-5.3-Codex, GPT-5.4-mini, and GPT-5.4 \citep{openai2026gpt53codex,openai2026gpt54,openai2026gpt54mini}.
The OpenAI models are evaluated with the Codex harness, while all remaining models use the Claude Code harness.
For models with reasoning-effort controls, we use the high setting throughout.

\paragraph{Metrics.}
We report three primary benchmark outcomes: feasibility, normalized quality score, and pass rate, following the scoring protocol in Section~\ref{sec:scoring-protocol}. 
We additionally record token usage, API cost, trajectory runtime as diagnostics for separating answer quality from agent efficiency.

\paragraph{Experimental settings.}
All evaluations are conducted in Harbor-based isolated environments without internet access. 
The task image provides a common optimization stack, including SCIP\citep{achterberg2009scip}. 
Each agent is given a 45-minute interaction budget, after which its submitted solution code is run under a 5-minute solve limit. 
For models supporting reasoning-effort controls, we use the high setting. 
All agents are initialized with the same four default OR-base skills.
% Experiments are run on an AMD Ryzen 9 9900X server with 24 threads and 123 GiB RAM. 
% Each task container is limited to 4 CPU cores and 8 GiB RAM.

\subsection{RQ1: How Well Do Frontier Coding Agents Perform?}
\label{sec:rq1}

The main experiment evaluates complete model-agent stacks on the full benchmark. As shown in Table~\ref{tab:main-results}, current agents exhibit a clear but limited frontier: the best stack passes 35.51\% of tasks, all other completed stacks stay below 35\%, and the best hard-task pass rate is only 20.59\%. Moreover, feasibility is consistently higher than pass rate, showing that agents often produce valid submissions without reaching sufficient optimization quality. This pattern reveals a central feasibility--quality gap in ORAgentBench. Current agents can construct acceptable decision artifacts, but remain structurally fragile when high-quality optimization requires stronger formulation and solving choices.

\begin{table}[t]
    \centering
    \caption{Main ORAgentBench results over 107 tasks, reported by difficulty split. The DeepSeek-V4-Pro row is averaged over three independent runs; other rows report one complete run. Bold and underline mark the best and second-best completed model-agent rows.}
    \label{tab:main-results}

    \begingroup
    \small
    \renewcommand{\arraystretch}{1.06}
    \setlength{\tabcolsep}{2.5pt}

    \begin{orbenchcompact}
    \makebox[\linewidth][c]{%
    \resizebox{\linewidth}{!}{%
    \begin{tabular}{@{}l DDD !{\color{ORBenchSubtle}\vrule width 0.35pt} DDD !{\color{ORBenchSubtle}\vrule width 0.35pt} DDDD@{}}
        \toprule
        \multirow{2}{*}{\textbf{Agent}} &
        \multicolumn{3}{c}{\textbf{Feasibility $\uparrow$}} &
        \multicolumn{3}{c}{\textbf{Quality $\uparrow$}} &
        \multicolumn{4}{c}{\textbf{Pass Rate (\%) $\uparrow$}} \\
        \cmidrule(lr){2-4}\cmidrule(lr){5-7}\cmidrule(l){8-11}
        & \multicolumn{1}{E}{\orcolhead{Easy}}
        & \multicolumn{1}{E}{\orcolhead{Medium}}
        & \multicolumn{1}{E}{\orcolhead{Hard}}
        & \multicolumn{1}{!{\color{ORBenchSubtle}\vrule width 0.35pt}E}{\orcolhead{Easy}}
        & \multicolumn{1}{E}{\orcolhead{Medium}}
        & \multicolumn{1}{E}{\orcolhead{Hard}}
        & \multicolumn{1}{!{\color{ORBenchSubtle}\vrule width 0.35pt}E}{\orcolhead{Easy}}
        & \multicolumn{1}{E}{\orcolhead{Medium}}
        & \multicolumn{1}{E}{\orcolhead{Hard}}
        & \multicolumn{1}{E}{\orcolhead{All}} \\
        \midrule

        \orgroupband{11}{Harness: \oragent{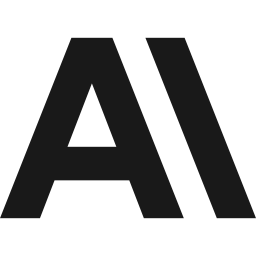}{Claude Code}} \\

        \oragent{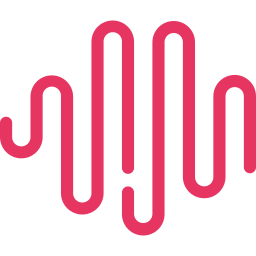}{MiniMax-M2.7} &
        0.50 & 0.20 & 0.06 &
        0.39 & 0.03 & 0.03 &
        40.62 & 2.44 & 2.94 & 14.02 \\

        \orstripe
        \oragent{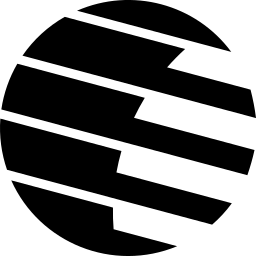}{Kimi K2.6} &
        0.50 & 0.51 & \orsecond{0.26} &
        0.29 & 0.25 & 0.13 &
        31.25 & 36.59 & 11.76 & 27.10 \\

        \oragent{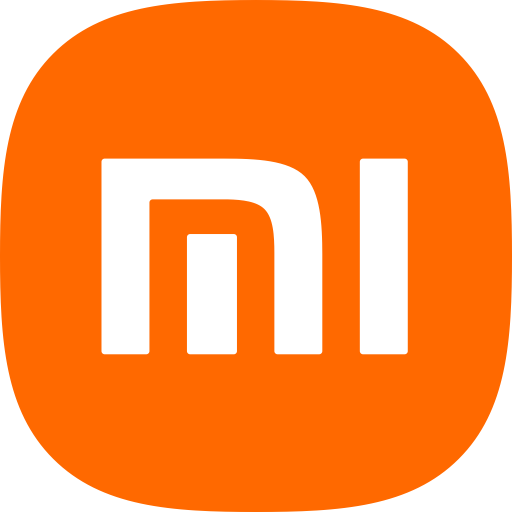}{MiMo V2.5 Pro} &
        0.59 & 0.46 & 0.06 &
        0.42 & 0.20 & 0.04 &
        43.75 & 21.95 & 2.94 & 22.43 \\

        \orstripe
        \oragent{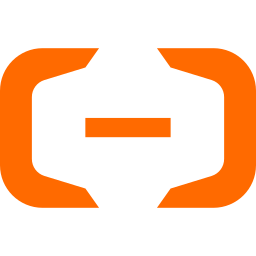}{Qwen3.5 Plus} &
        0.59 & 0.24 & 0.03 &
        0.45 & 0.13 & 0.03 &
        46.88 & 14.63 & 2.94 & 20.56 \\

        \oragent{alibabacloud}{Qwen3.6 Plus} &
        0.50 & 0.29 & 0.15 &
        0.38 & 0.21 & 0.07 &
        40.62 & 21.95 & 5.88 & 22.43 \\

        \orstripe
        \oragent{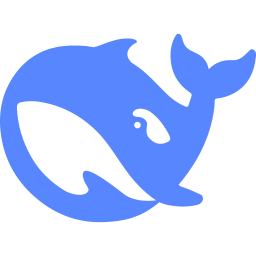}{DeepSeek V4 Flash} &
        0.62 & 0.41 & 0.15 &
        0.45 & 0.21 & 0.09 &
        46.88 & 26.83 & 5.88 & 26.17 \\

        \oragent{deepseek}{DeepSeek V4 Pro} &
        0.55 & 0.46 & 0.23 &
        0.38 & 0.25 & 0.14 &
        39.58 & 28.46 & 13.73 & 27.10 \\

        \orstripe
        \orzhipuagent{GLM 5} &
        0.53 & 0.34 & 0.12 &
        0.35 & 0.21 & 0.06 &
        37.50 & 24.39 & 5.88 & 22.43 \\

        \orzhipuagent{GLM 5.1} &
        0.53 & 0.46 & 0.15 &
        0.37 & 0.29 & 0.10 &
        37.50 & 31.71 & 8.82 & 26.17 \\

        \orstripe
        \orclaudemodelagent{Claude Sonnet 4.6} &
        0.47 & 0.51 & \orsecond{0.26} &
        0.31 & 0.23 & 0.14 &
        31.25 & 29.27 & \orsecond{14.71} & 25.23 \\

        \orclaudemodelagent{Claude Opus 4.6} &
        \orsecond{0.66} & 0.59 & 0.21 &
        \orbest{0.49} & \orsecond{0.32} & 0.14 &
        \orbest{50.00} & \orsecond{39.02} & \orsecond{14.71} & \orsecond{34.58} \\

        \addlinespace[1pt]
        \orgroupband{11}{Harness: \oropenaiagent{Codex}} \\

        \orstripe
        \oropenaiagent{GPT-5.3-Codex} &
        0.50 & \orbest{0.68} & \orbest{0.32} &
        0.30 & \orbest{0.37} & \orsecond{0.19} &
        31.25 & \orbest{43.90} & \orbest{20.59} & 32.71 \\

        \oropenaiagent{GPT-5.4-mini} &
        0.62 & 0.54 & 0.21 &
        0.43 & 0.24 & 0.07 &
        43.75 & 29.27 & 5.88 & 26.17 \\

        \orstripe
        \orbest{\oropenaiagent{GPT-5.4}} &
        \orbest{0.69} & \orbest{0.68} & \orbest{0.32} &
        \orsecond{0.45} & 0.31 & \orbest{0.21} &
        \orsecond{46.88} & \orsecond{39.02} & \orbest{20.59} & \orbest{35.51} \\

        \bottomrule
    \end{tabular}%
    }%
    }
    \end{orbenchcompact}

    \endgroup
\end{table}

% \ortakeaway{1}{ORAgentBench exposes a gap beyond executable code: current agents often reach feasibility, but still fail to produce high-quality operational decisions.}

\begin{figure}[thbp]
    \centering
    \includegraphics[width=\linewidth]{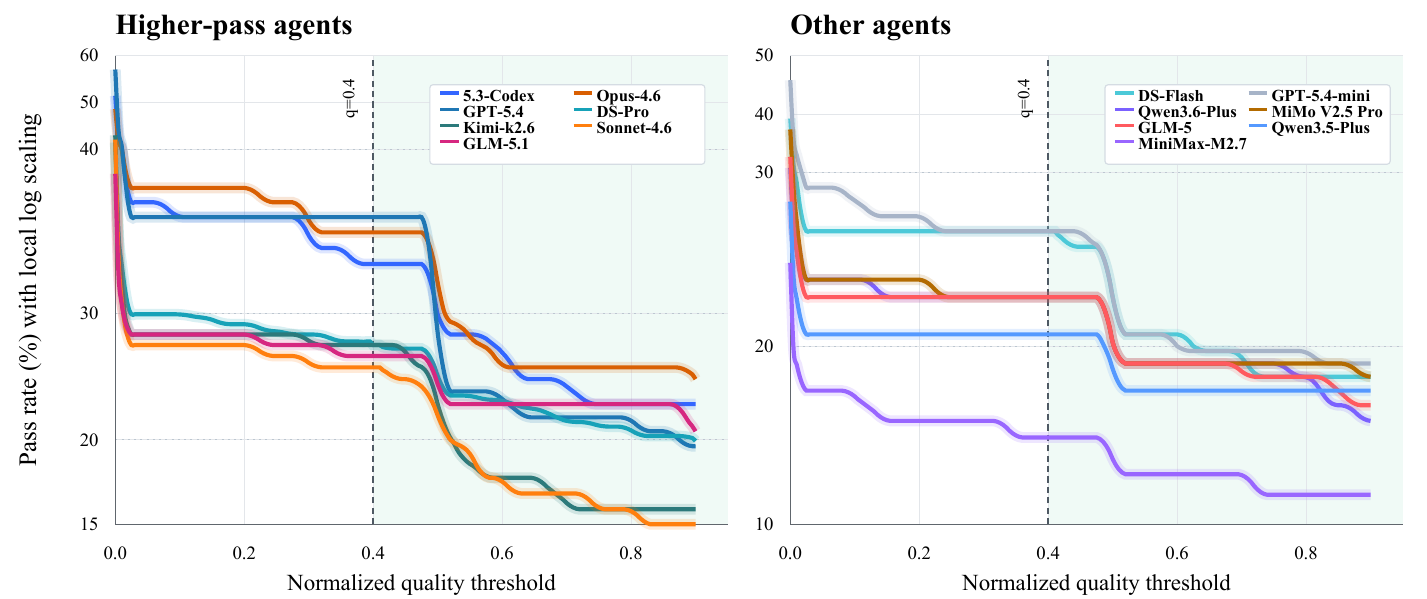}
    \caption{Pass rate under increasing quality thresholds, showing sensitivity to solution quality beyond feasibility. Curves are split by all-task pass tier for readability.}
    \label{fig:threshold-curve}
\end{figure}

Figure~\ref{fig:threshold-curve} examines how pass rates change as the quality threshold increases. Around the default threshold $q=0.4$, several agents have similar pass rates, but their curves diverge under stricter thresholds. 
Agents whose curves drop sharply mainly produce solutions that barely satisfy the default criterion, whereas stronger agents decline more gradually and retain nontrivial pass rates at higher quality levels. This shows that ORAgentBench is sensitive not only to feasibility, but also to solution-quality margins. 
It therefore distinguishes agents that merely find acceptable feasible decisions from those that produce more robust, higher-quality optimization decisions.

\subsection{RQ2: Efficiency, Runtime, and Cost}
\label{sec:efficiency}

\begin{figure}[htbp]
    \centering
    \includegraphics[width=\linewidth]{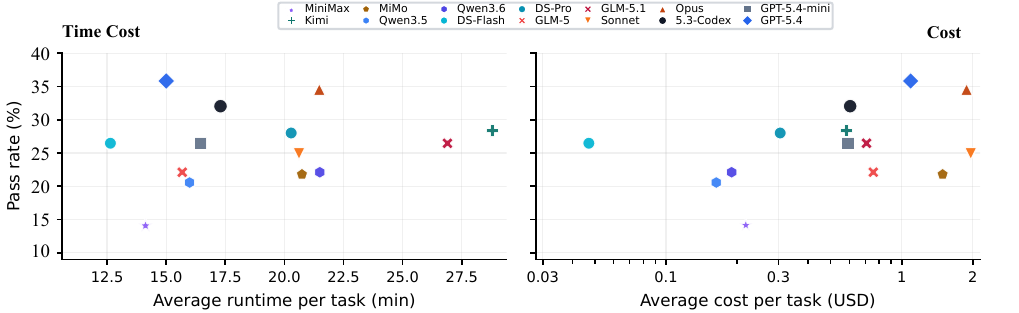}
    \caption{Efficiency trade-off across model-agent rows. Left: pass rate versus average runtime per task. Right: pass rate versus average API cost per task in USD.}
    \label{fig:rq2-efficiency-cost-runtime}
\end{figure}

ORAgentBench also exposes efficiency trade-offs. Figure~\ref{fig:rq2-efficiency-cost-runtime} shows that pass rate is not monotonic with either average runtime or API cost. 
The best-performing stack is neither the longest-running nor the most expensive, and several costly or slow stacks do not achieve higher pass rates.
This suggests that additional computation is useful only when it is spent on the right stages, such as data checking, feasible model construction, and targeted solution improvement. 
Simply extending trajectories or increasing token budget does not guarantee better OR decisions.
Overall, ORAgentBench highlights an important efficiency--quality trade-off: strong OR agents require not only more resources, but better allocation of reasoning and solving effort.

% \ortakeaway{2}{Efficiency in OR agents is about disciplined trajectory allocation, not simply lower cost or longer reasoning.}

\subsection{RQ3: What Are the Dominant Modeling and Solving Failures?}
\label{sec:rq3}

\noindent
% preamble 中需要：
% \usepackage{capt-of}   % for \captionof

\noindent
\begin{minipage}[t]{0.50\linewidth}
    \vspace{0pt}
\textbf{Failure-mode diagnostics.} 
We diagnose failed trajectories by the first stage at which they break. 
\emph{Modeling errors} refer to missing, inconsistent, or incorrectly encoded operational rules; \emph{formulation efficiency} denotes plausible models that become too large or brittle to yield a valid artifact. 
\emph{Weak solver} captures feasible submissions that fail to meet the quality threshold, while \emph{timeout} indicates trajectories that exhaust the budget without preserving an accepted fallback. Figure~\ref{fig:failure_distribution} shows that the dominant source of failure is not infrastructure or raw solver time, but strategy. 
Modeling-side failures account for 54.8\% of non-passing trials, 
\end{minipage}
\hfill
\begin{minipage}[t]{0.48\linewidth}
    \vspace{-0.5\baselineskip}
    \centering
    \includegraphics[width=\linewidth]{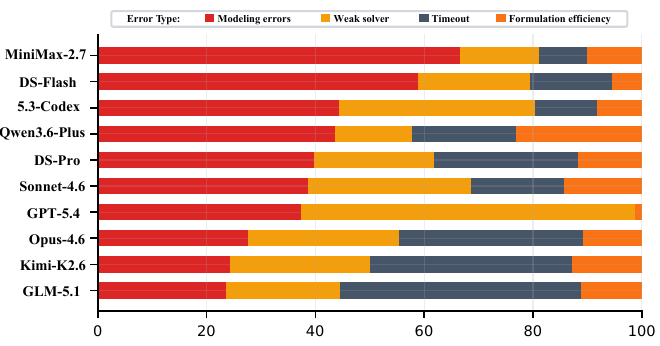}
    \captionof{figure}{Failure-mode diagnostics across model-agent rows, grouped by the first failed stage in each non-passing trajectory.}
    \label{fig:failure_distribution}
\end{minipage}

\vspace{-0.2\baselineskip}
\noindent
indicating that many agents fail before solver strength can help. 
The model-specific breakdown is also informative: some agents mainly miss operational rules, others spend the budget without safeguarding feasibility, and stronger agents more often shift the bottleneck toward solution improvement. 
This pattern suggests that simply giving agents more solve time is unlikely to close the gap; progress requires better modeling strategies, fallback mechanisms, and targeted improvement procedures.

% \ortakeaway{3}{Most failures originate before solver strength becomes decisive: agents need better rule recovery, fallback preservation, and improvement policies.}

\begin{figure}[t]
    \centering
    \includegraphics[width=\linewidth]{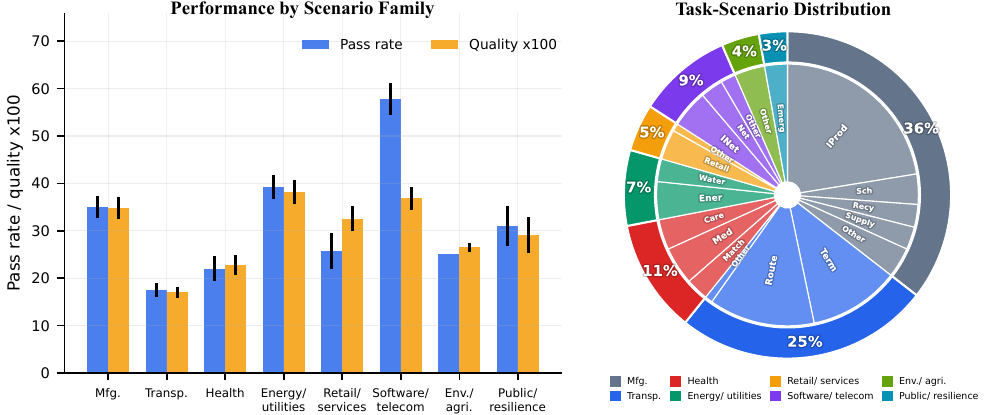}
    \caption{Problem-family diagnostics. Left: pass rate and normalized quality by scenario family, with standard-error bars across model-agent rows. Right: Task-Scenario Distribution, showing coarse family proportions and abbreviated subfamily labels.}
    \label{fig:family-results}
    \vspace{-1.0em}
\end{figure}

\noindent
\textbf{Problem-family diagnostics.}
Figure~\ref{fig:family-results} reports performance across eight scenario families and the corresponding Task-Scenario Distribution: manufacturing (Mfg.), transportation (Transp.), healthcare (Health), energy/utilities, retail/services, software/telecom, environment/agriculture (Env./agri.), and public/resilience. 
The results show substantial cross-family variation: transportation is among the hardest families, while software/telecom achieves much higher pass rates. 
This variation is not explained by the domain label alone, but by the modeling structures induced by each setting. 
Tasks with continuity, recourse, sparse feasible regions, or multi-period coupling make both feasibility and quality improvement more fragile. 
By contrast, families with more explicit graph, capacity, or coverage structures are generally easier, although quality still drops when substitution or survivability constraints become central. 
Overall, the scenario-family results show that ORAgentBench captures hidden modeling burden behind surface application categories.

% \ortakeaway{4}{Scenario labels are useful only when tied to structure: continuity, recourse, sparse feasibility, and coupling drive the observed performance gaps.}

\subsection{RQ4: Do OR Skills Help?}
\label{sec:rq4-skills}

The skill-enhancement study examines whether procedural OR guidance improves agent behavior.
We fix the agent stack to Codex + GPT-5.4 and compare three settings: \textit{Without Skills}, which provides no external guidance; \textit{Base Skills}, which uses the generic skills supplied in the main benchmark; and \textit{Expert Skills}, which uses high-quality OR skills manually written by domain experts.
The results show that skills reshape the failure profile rather than uniformly improving performance.
Base Skills mainly improve feasibility, but these gains do not consistently translate into higher quality or pass rate: easy tasks degrade, while medium and hard tasks see only modest pass-rate gains.
Expert Skills show a stronger difficulty-dependent effect, improving hard-task feasibility and pass rate but reducing performance on easier tasks.
This suggests that OR-specific guidance is most useful when tasks require feasibility protection, structured search, and fallback reasoning, but can add overhead or misaligned guidance for simpler instances.
As detailed in Appendix~\ref{app:expert-skills}, the paired analysis confirms this trade-off: Expert Skills recover some infeasible runs but also introduce new failures, including more pass-to-fail conversions than fail-to-pass gains.
Trajectory evidence further shows that skills help only when they trigger task-specific diagnosis, search, and revision; procedural compliance alone can increase internal confidence without improving correctness.
Since each condition contains one complete run, these findings should be interpreted as descriptive evidence rather than a variance-controlled causal estimate.

% The skills encode workflow discipline: audit the data, construct and validate a feasible model, and then improve solution quality.
% This targets the dominant failure modes above, including missed constraints, brittle formulations, and weak fallback strategies.

\begin{table}[H]
    \centering
    \vspace{-0.6em}
    \caption{Skill-enhancement analysis. Colored deltas are absolute changes from Without Skills.}
    \label{tab:skill-results}
    \begin{orbenchcompact}
    \scriptsize
    \setlength{\tabcolsep}{2pt}
    \newcommand{\updelta}[1]{\textcolor{green!45!black}{\tiny(+#1)}}
    \newcommand{\downdelta}[1]{\textcolor{red!65!black}{\tiny(-#1)}}
    \newcommand{\sd}[2]{\shortstack{#1\\#2}}
    \begin{tabular*}{\linewidth}{@{\extracolsep{\fill}}llccc ccc ccc@{}}
        \toprule
        \multirow{2}{*}{\textbf{Setting}} 
        & \multirow{2}{*}{\textbf{Condition}} 
        & \multicolumn{3}{c}{\textbf{Feasibility $\uparrow$}} 
        & \multicolumn{3}{c}{\textbf{Quality $\uparrow$}} 
        & \multicolumn{3}{c}{\textbf{Pass Rate(\%) $\uparrow$}} \\
        \cmidrule(lr){3-5} \cmidrule(lr){6-8} \cmidrule(l){9-11}
        & & \textbf{Easy} & \textbf{Medium} & \textbf{Hard}
          & \textbf{Easy} & \textbf{Medium} & \textbf{Hard}
          & \textbf{Easy} & \textbf{Medium} & \textbf{Hard} \\
        \midrule
        \orstripe
        Without Skills
        & No skill
        & 0.81 & 0.66 & 0.24
        & 0.61 & 0.35 & 0.20
        & 62.50 & 39.02 & 17.65 \\
        \midrule
        \multirow{2}{*}{With Skills}
        & Base Skills
        & \sd{0.69}{\downdelta{0.12}} & \sd{0.68}{\updelta{0.02}} & \sd{0.32}{\updelta{0.08}}
        & \sd{0.45}{\downdelta{0.16}} & \sd{0.31}{\downdelta{0.04}} & \sd{0.21}{\updelta{0.01}}
        & \sd{46.88}{\downdelta{15.62}} & \sd{39.02}{\updelta{0.00}} & \sd{20.59}{\updelta{2.94}} \\
        & Expert Skills
        & \sd{0.84}{\updelta{0.03}} & \sd{0.59}{\downdelta{0.07}} & \sd{0.35}{\updelta{0.11}}
        & \sd{0.56}{\downdelta{0.05}} & \sd{0.29}{\downdelta{0.06}} & \sd{0.23}{\updelta{0.03}}
        & \sd{59.81}{\downdelta{2.69}} & \sd{36.15}{\downdelta{2.87}} & \sd{21.50}{\updelta{3.85}} \\
        \bottomrule
    \end{tabular*}
    \end{orbenchcompact}
\end{table}

\section{Conclusion}
\label{conclusion}

We introduced ORAgentBench, an execution-grounded benchmark for evaluating agents on realistic end-to-end operations research workflows. Across 107 isolated tasks, agents must interpret operational artifacts, design modeling and solving strategies, implement executable methods, and submit decisions judged by feasibility and objective quality. Our results show that frontier coding agents can solve some OR tasks, but remain far from reliable autonomous OR practice: many failures arise from missing business rules, brittle formulations, weak feasible-solution construction, and poor improvement strategies rather than from coding alone. The six-dimensional construction rubric further shows that OR difficulty is not reducible to scale; formulation structure, constraint coupling, dynamic state, data size, and problem understanding stress different agents in different ways. ORAgentBench therefore provides both a benchmark and a diagnostic framework for developing agents that can move beyond plausible optimization code toward dependable, high-quality operational decisions.

% \newpage
\medskip
\bibliographystyle{plainnat}
\bibliography{ref}

%%%%%%%%%%%%%%%%%%%%%%%%%%%%%%%%%%%%%%%%%%%%%%%%%%%%%%%%%%%%
\newpage
\appendix

\section{Limitations and Future Directions.}
ORAgentBench evaluates executable optimization workflows, but it does not cover the full scope of OR practice.
Real OR projects often involve stakeholder negotiation, qualitative policy judgment, live feedback, and long-term deployment, which are beyond deterministic validation.
Our single-step tasks use fixed operational snapshots, and our multi-step tasks use finite scripted events; neither fully captures open-ended distribution shifts, evolving business rules, or all data-quality failures.
Our evaluation also focuses on the final submitted solution.
Although ORAgentBench records full trajectories and generated artifacts, it does not yet provide a complete process-level audit of agent behavior.
Future benchmarks should evaluate not only feasibility and solution quality, but also intermediate reasoning, data understanding, formulation choices, solver interaction, debugging, and self-validation.
Finally, ORAgentBench is designed for evaluation rather than training.
A key future direction is to build scalable OR environments from diverse real-world scenarios, enabling agents to learn how to explore files, diagnose failures, adapt solving strategies, and improve through feedback.
Connecting benchmark evaluation with environment-based training may be crucial for building OR agents that generalize to realistic industrial settings.

\section{Additional Benchmark Details}
\label{app:details}

\subsection{Additional Scoring Details}
\label{app:scoring-details}

The Harbor verifier first evaluates schema and hard constraints. For a single-step task, failure sets feasibility and quality to zero. For a feasible submission, let $O$ be the submitted objective, $R$ the verified reference objective, and $B$ a valid global bound. Define the signed improvement over the reference and the reference-to-bound width as
\[
g(O)=
\begin{cases}
R-O, & \text{minimization},\\
O-R, & \text{maximization},
\end{cases}
\qquad
W=
\begin{cases}
R-B, & \text{minimization},\\
B-R, & \text{maximization}.
\end{cases}
\]
When $B$ is valid and $W$ is numerically meaningful, the raw quality score is
\[
Q_{\mathrm{raw}}=\operatorname{clip}\!\left(1+\frac{g(O)}{W},0,2\right).
\]
Thus, a reference-level solution receives the midpoint score, a solution approaching the valid bound approaches the maximum, and sufficiently poor feasible solutions approach zero. If the bound and reference are indistinguishable at the task tolerance, we instead use
\[
d=\frac{g(O)}{\max(|R|,10^{-8})\,\epsilon_\tau},\qquad
Q_{\mathrm{raw}}=
\begin{cases}
\operatorname{clip}(2+d,0,2), & d\geq -1,\\
\operatorname{clip}(1.5+0.5d,0,2), & d<-1,
\end{cases}
\]
where $\epsilon_\tau$ is the task's accepted relative-gap tolerance. When no defensible global bound exists, the evaluator uses the verified reference as the anchor and applies the task's recorded tolerance or reference-to-poor calibration; all variants return $Q_{\mathrm{raw}}\in[0,2]$ and clip improvements at the maximum.

The normalized quality reported in the paper is $q=Q_{\mathrm{raw}}/2\in[0,1]$. The scalar Harbor reward stored in each trial is
\[
r=
\begin{cases}
(1+Q_{\mathrm{raw}})/3, & F=1,\\
0, & F=0,
\end{cases}
\]
where $F\in\{0,1\}$ is feasibility. The binary pass indicator used in all tables is
\[
\mathrm{Pass}=\mathbb{I}[F=1\land q>0.4].
\]
We keep these three quantities separate: $q$ is the paper's normalized quality, $r$ is Harbor's scalar reward, and Pass is the thresholded benchmark outcome. In a multi-step task, stage qualities are retained for diagnosis and aggregated according to \texttt{task.toml}; overall feasibility is strict across the required trajectory. Hence an infeasible stage makes $F=0$, $r=0$, and Pass false even when earlier stage qualities were high.

\subsection{Seed Data Source}
\label{app:seed-sources}

We audited all 107 task packages using their provenance records and construction notes. Table~\ref{tab:seed-sources} reports the primary seed that supplied the operational setting or mathematical core. Each task is counted once under this primary origin; secondary data, regulations, and synthetic augmentation are recorded but not double-counted.

\begin{table}[htbp]
\centering
\caption{Primary construction sources of ORAgentBench tasks.}
\label{tab:seed-sources}
\begin{orbenchtable}
\begin{tabularx}{0.88\linewidth}{@{}Xrr@{}}
\toprule
\textbf{Primary source} & \textbf{Tasks} & \textbf{Share} \\
\midrule
OR papers and application studies & 49 & 45.8\% \\
\orstripe
IndustryOR benchmark seeds & 29 & 27.1\% \\
Operational scenarios & 19 & 17.8\% \\
\orstripe
Public datasets and repositories & 10 & 9.3\% \\
\midrule
\textbf{Total} & \textbf{107} & \textbf{100.0\%} \\
\bottomrule
\end{tabularx}
\end{orbenchtable}
\end{table}

For paper- and application-derived tasks, we retain the decision context and principal OR structure, then construct a new fixed instance with calibrated scale, interacting rules, and an auditable submission schema. IndustryOR seeds are expanded from text-to-formulation problems into executable tasks with structured data and hidden validation. Public datasets are cleaned and frozen at a documented snapshot, while operational scenarios and legacy packages are converted into self-contained artifacts. Construction records preserve source identifiers, snapshot notes, transformation summaries, and the role of synthetic fields where applicable. Synthetic augmentation completes missing fields, creates cross-file dependencies, or calibrates feasibility and difficulty; it does not copy a reference solution into the public packet. The benchmark therefore tests transfer from a source setting rather than reproduction of a published solution.

\subsection{Task Artifacts}
\label{app:task-artifacts}

Each task is a self-contained Harbor package. The package separates the public decision environment from the private grading layer:

\begin{orappendixbox}{Single-step Harbor task}
\begin{lstlisting}[style=orbenchcode]
harbor_tasks/<task>/
|-- instruction.md              # agent-facing workflow and deliverables
|-- task.toml                   # Harbor, Docker, resources, and time limits
|-- environment/
|   |-- Dockerfile              # builds the isolated runtime
|   |-- skills/                 # optional procedural guidance
|   `-- app/                    # mounted as /app; visible to the agent
|       |-- PROBLEM_STATEMENT.md
|       |-- data/               # public instance and configuration files
|       `-- submissions/        # templates and agent-written artifacts
|-- tests/                      # private evaluator and reference metrics
`-- solution/                   # private oracle/reference implementation
\end{lstlisting}
\end{orappendixbox}

The agent receives \texttt{instruction.md} and works in the image built from \texttt{environment/}. Within that image, \texttt{environment/app} becomes \texttt{/app} and contains the operational brief, public data, submission templates, and all task-specific information needed to solve the instance. When enabled, \texttt{environment/skills} is made available at \texttt{/skills} as generic procedural guidance. The agent writes its mathematical model, executable solver, solve log, and final decision artifact under \texttt{/app/submissions}.

The remaining components serve different roles. \texttt{task.toml} is Harbor orchestration metadata rather than part of the optimization instance: it specifies the container build, work directory, resources, time limits, retained artifacts, and, for multi-step tasks, step order and reward aggregation. The \texttt{tests} directory contains the private evaluator and reference metrics, while \texttt{solution} contains the private reference implementation used to establish grading targets. Neither is exposed as solving evidence. After execution, Harbor archives the declared submissions, agent trajectory, and verifier logs; these records support analysis but are not additional inputs available during the run.

\paragraph{Multi-step tasks.}
Online and dynamic tasks retain the same public/private boundary but organize each decision epoch under \texttt{steps/}. A later step is not a fresh independent instance: it updates the existing workspace while preserving the agent's earlier submissions and state.

\begin{orappendixbox}{Multi-step Harbor task}
\begin{lstlisting}[style=orbenchcode]
harbor_tasks/<task>/
|-- task.toml                   # ordered steps and reward aggregation
|-- environment/app/           # initial public workspace
|   |-- PROBLEM_STATEMENT.md
|   |-- EVENT_NOTICE.md
|   |-- data/{{CURRENT_STATE_FILE}}
|   `-- submissions/
`-- steps/
    |-- initial_plan/
    |   |-- instruction.md
    |   |-- tests/              # private step-specific evaluator
    |   `-- solution/           # private step-specific reference
    |-- step2_replan/
    |   |-- instruction.md
    |   |-- workdir/            # event notice, data update, setup
    |   |-- tests/
    |   `-- solution/
    `-- step3_final_replan/
        |-- instruction.md
        |-- workdir/
        |-- tests/
        `-- solution/
\end{lstlisting}
\end{orappendixbox}

\begin{center}
\begin{tikzpicture}[
  node distance=0.55cm,
  stage/.style={draw=ORCaseInfo, fill=ORCaseInfoBack, rounded corners=1pt,
    align=center, text width=0.25\linewidth, minimum height=1.25cm,
    font=\small},
  arrow/.style={-{Latex[length=2mm]}, thick, draw=ORCaseInfo}
]
\node[stage] (s1) {\textbf{Step 1: initial plan}\\Read the initial packet\\Submit the first plan};
\node[stage, right=of s1] (s2) {\textbf{Step 2: event update}\\Overlay new information\\Preserve frozen decisions};
\node[stage, right=of s2] (s3) {\textbf{Step 3: final replan}\\Repair the open horizon\\Submit the updated plan};
\draw[arrow] (s1) -- (s2);
\draw[arrow] (s2) -- (s3);
\end{tikzpicture}
\end{center}

At runtime, Harbor creates the environment once. The initial step uses the base workspace; before each later step, \texttt{workdir/setup.sh} overlays the newly revealed event notice and data while retaining the artifacts declared in \texttt{task.toml}, typically \texttt{/app/submissions}, \texttt{/app/state}, and a current-period or current-wave file. The new state may be computed from the previously accepted plan, so it represents operational consequences rather than a fresh independent instance. The step instruction names the planning horizon, prior baseline, frozen commitments, and required output files.

Each step has its own private evaluator and reference solution. Decisions inside a freeze window remain binding, while later commitments may be changed only under the stated replanning rules and penalties. A minimum-reward gate controls whether the next event is revealed. Stage scores diagnose local progress, but the paper-level outcome is computed from the complete required trajectory: every required stage must remain feasible, and any infeasible replan makes task-level feasibility zero. Multi-step tasks therefore test state persistence, information discipline, and controlled repair in addition to one-shot modeling and optimization.

\section{Additional Experiment Analysis}
\label{app:additional-analysis}

\subsection{Difficulty-Dimension Diagnostics}

\begin{figure}[H]
    \centering
    \includegraphics[width=\linewidth]{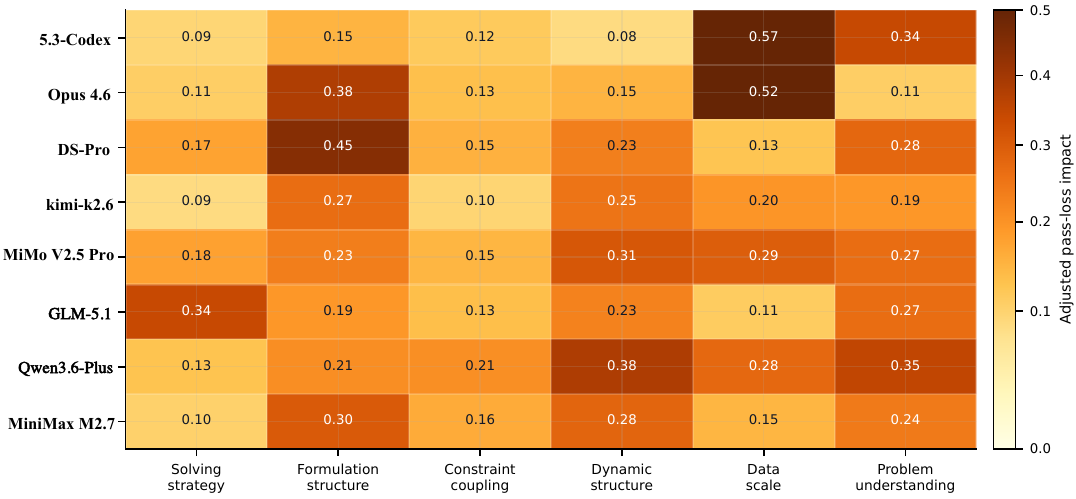}
    \caption{Difficulty attribution by model. Cells report the predicted increase in pass loss when a construction dimension moves from 0 to 3; darker values indicate larger associations.}
    \label{fig:difficulty-attribution}
    \vspace{-1.0em}
\end{figure}

We relate observed failures to the six construction-time dimensions in Section~\ref{sec:design-principles}. To avoid a crowded diagnostic, Figure~\ref{fig:difficulty-attribution} shows one representative row from each of eight model families; the main table remains the complete comparison. For each displayed row and all 107 tasks, define $D_i=1-\mathrm{Pass}_i$. We fit the non-negative multivariate model
\[
D_i=\alpha+\sum_{d=1}^{6}\beta^{\mathrm{multi}}_d x_{id}+\epsilon_i,
\]
and shrink each coefficient toward its non-negative marginal slope:
\[
\beta_d=0.78\beta^{\mathrm{multi}}_d+0.22\beta^{\mathrm{marg}}_d,
\qquad
\mathrm{AdjustedImpact}_d=3\beta_d.
\]
The multivariate coefficients are obtained by non-negative active-set fitting. The fixed $0.78/0.22$ blend is applied uniformly to stabilize correlated dimensions, and values below a small numerical display floor remain visually distinguishable in the heatmap. The reported impact is the predicted increase in pass loss when a dimension moves from 0 to 3. This is a descriptive attribution, not a causal estimate or a model ranking.

Difficulty is not explained by scale alone. Formulation structure has the most consistent association with pass loss, while data scale and dynamic structure affect model families differently. The smaller joint coefficient for constraint coupling partly reflects its correlation with formulation and dynamic structure; trajectory analysis nevertheless identifies coupled rules as a frequent source of invalid submissions.

\subsection{Solution-Method Analysis}
\label{app:solution-methods}

\begin{figure}[htbp]
    \centering
    \includegraphics[width=\linewidth]{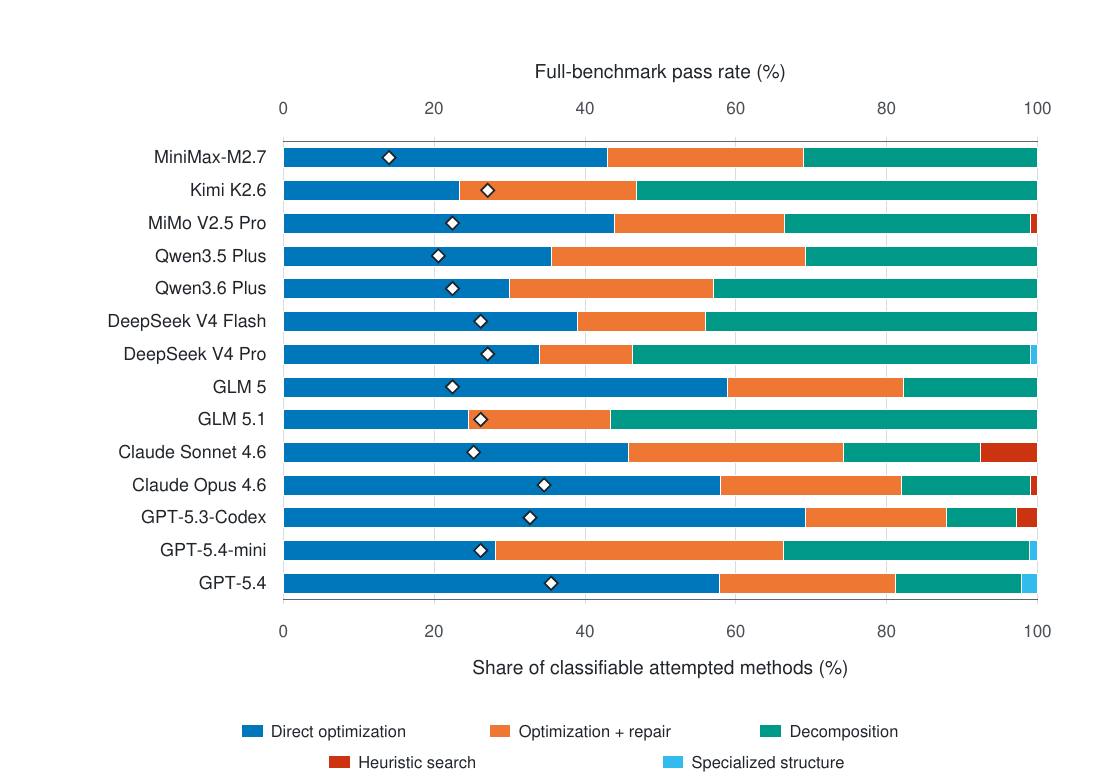}
    \caption{Solution methods in all trajectories. Stacked bars show method shares among classifiable, deduplicated model-task records. Open diamonds show the independent full-benchmark pass rates from Table~\ref{tab:main-results}.}
    \label{fig:solution-methods}
\end{figure}

We code the dominant method \emph{attempted} in each retained model--task record, regardless of whether the run ultimately produced a feasible or high-quality solution. The coding is based on agent-authored reasoning, tool actions, solver code, and solve logs from all retained steps, while task instructions and skill text are excluded to avoid conflating provided guidance with the agent's actual strategy. Repeated backup attempts are deduplicated at the model--task level. Figure~\ref{fig:solution-methods} reports method shares over classifiable attempts.

\paragraph{Method categories.} \emph{Direct optimization} formulates the task as an LP or MIP and works primarily through the solver, including runs that remain in formulation or solver debugging. \emph{Optimization + repair} combines an exact model with a constructed incumbent, warm start, variable fixing, restricted re-optimization, rounding, or local repair. \emph{Decomposition} separates a large problem by horizon, scenario, route set, or decision stage and coordinates smaller master and subproblems, including rolling-horizon procedures. \emph{Heuristic search} covers constructive and greedy rules as well as neighborhood or metaheuristic improvement when a full exact model is impractical. \emph{Specialized structure} uses algorithms matched to a particular representation, such as network flow, matching, shortest path, or constraint programming. When several signals occur, we assign the approach that governs the implemented search rather than counting one run multiple times.

These bars describe attempted strategy, not method success: debugging-only runs remain counted under the method they implemented. Direct optimization dominates several agents, whereas others more often introduce decomposition or repair. The same labels occur in passing and failing runs, so the figure does not support a causal comparison of methods. Across trajectories, the more consistent distinction is procedural: successful runs match the method to task structure, preserve a feasible incumbent during improvement, and validate the exported artifact under the stated objective.

\subsection{Run-to-Run Variability}
\label{app:deepseek-cross-job-variability}

We quantify the stochastic run-to-run variability of the DeepSeek-V4-Pro + Claude Code configuration using three evaluations under the Without Skills condition: two repeated trials from one job and the retained complete run closest to the main-table profile. All three runs use the same task set, model backend, agent harness, skill setting, and execution protocol. This analysis is intended to estimate the empirical variability induced by repeated agent runs under an identical configuration, rather than to compare different skill settings.

\begin{figure}[htbp]
\centering
\includegraphics[width=0.88\linewidth]{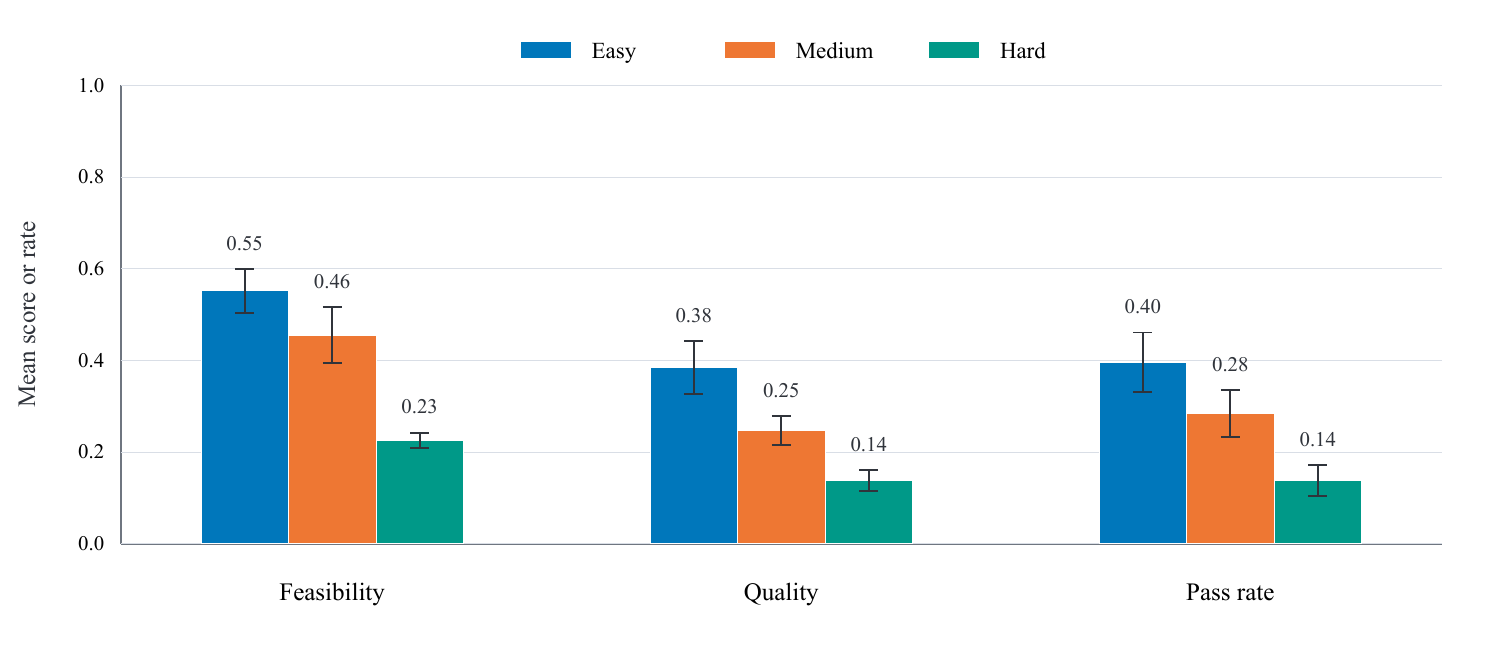}
\caption{Run-to-run variability of DeepSeek-V4-Pro + Claude Code across three Without Skills runs. Each metric group contains Easy, Medium, and Hard bars; bar heights show the three-run mean, and error bars show the sample standard deviation.}
\label{fig:deepseek-cross-job-variability}
\end{figure}

Figure~\ref{fig:deepseek-cross-job-variability} summarizes the three repeated runs. For each metric and difficulty band, bars show the mean across runs, and error bars show the sample standard deviation. Pass rate is plotted as a proportion so that feasibility, quality, and pass rate share the same $[0,1]$ scale.

The repeated runs show a consistent difficulty trend: performance is highest on Easy tasks and declines substantially on Medium and Hard tasks. Dispersion is modest: the largest sample standard deviation is $0.065$ for Easy-task pass rate, and Hard-task feasibility remains low with a sample standard deviation of $0.017$. These results indicate that stochastic variation exists even under the same model--harness--skill configuration, but the magnitude of this variation does not change the main difficulty pattern. Since only three repetitions are available, we report sample standard deviations as descriptive estimates of run-to-run variability rather than confidence intervals.

Treating the three outcomes for each task as one matched group, Table~\ref{tab:deepseek-pass3} reports the mean single-run pass rate, $\mathrm{pass@3}$ (at least one successful run), and $\mathrm{pass^3}$ (three successful runs). The corresponding success counts are 18/32, 18/41, and 11/34 for $\mathrm{pass@3}$, versus 7/32, 6/41, and 1/34 for $\mathrm{pass^3}$.

\begin{table}[htbp]
\centering
\caption{Matched-task pass rates across three DeepSeek-V4-Pro runs. Colored
parentheses report the difference from the mean pass rate in percentage points.}
\label{tab:deepseek-pass3}
\begin{orbenchtable}
\begin{tabular}{@{}lccc@{}}
\toprule
& \textbf{Easy} & \textbf{Medium} & \textbf{Hard} \\
\midrule
Mean pass rate &
39.6\% {\scriptsize\textcolor{ORBenchSubtle}{(0.0 pp)}} &
28.5\% {\scriptsize\textcolor{ORBenchSubtle}{(0.0 pp)}} &
13.7\% {\scriptsize\textcolor{ORBenchSubtle}{(0.0 pp)}} \\
$\mathrm{pass@3}$ &
56.3\% {\scriptsize\textcolor{ORCaseSuccess}{(+16.7 pp)}} &
43.9\% {\scriptsize\textcolor{ORCaseSuccess}{(+15.4 pp)}} &
32.4\% {\scriptsize\textcolor{ORCaseSuccess}{(+18.6 pp)}} \\
$\mathrm{pass^3}$ &
21.9\% {\scriptsize\textcolor{ORCaseFail}{(-17.7 pp)}} &
14.6\% {\scriptsize\textcolor{ORCaseFail}{(-13.8 pp)}} &
2.9\% {\scriptsize\textcolor{ORCaseFail}{(-10.8 pp)}} \\
\bottomrule
\end{tabular}
\end{orbenchtable}
\end{table}

\section{Skills}
\label{app:skills}

\subsection{Base Skills}

The Base Skills collection contains four general-purpose skills for formulation, implementation, and repair. It is stored under \nolinkurl{ORAgentBench/skills/base-Skills} and is the default skill collection used in the main benchmark. Each skill is an agent-readable \texttt{SKILL.md} file mounted under \texttt{/skills}; it supplies a reusable workflow rather than executable task logic. The skills contain no task data, evaluator rules, reference objectives, or solutions, so the agent must still inspect the public packet, formulate the optimization problem, implement the solver, and validate the submitted artifact. Table~\ref{tab:base-skills} lists every Base Skill using the name and description declared in its \texttt{SKILL.md} header.

\begin{table}[H]
\centering
\caption{Base skills available to evaluated agents.}
\label{tab:base-skills}
\begin{orbenchtable}
\begin{tabularx}{\linewidth}{@{}p{0.31\linewidth}Y@{}}
\toprule
\textbf{Skill} & \textbf{Description} \\
\midrule
\texttt{advanced-mip-patterns} & Use when a task involves fixed charges, logical implications, min/max choices, sequencing, no-overlap, piecewise costs, routing-state logic, or other advanced MIP modeling patterns. \\
\orstripe
\texttt{optimization-modeling-core} & Use for natural-language-to-optimization tasks that require translating a business problem into a complete mathematical model before writing solver code. \\
\texttt{optimization-solver-repair} & Use when optimization solver code crashes, produces no solution, is rejected by a checker, or returns infeasible/unbounded; guides repair by distinguishing code errors from modeling errors and keeping the written model and code synchronized. \\
\orstripe
\texttt{or-pyscipopt} & Use when building, running, validating, or debugging Python optimization models that use PySCIPOpt and SCIP. \\
\bottomrule
\end{tabularx}
\end{orbenchtable}
\end{table}

\subsection{Expert Skills}
\label{app:expert-skills}

The Expert Skills collection is stored under \nolinkurl{ORAgentBench/skills/Expert-Skills}. It contains nine expert skills covering the general OR workflow, decomposition, constraint programming, metaheuristics, multi-objective and stochastic optimization, network flow, routing, and evidence-driven diagnosis. These skills provide method-selection criteria, formulation patterns, implementation guidance, and independent validation procedures for the corresponding OR structures. They remain task-independent and do not expose benchmark instances or evaluator information. Table~\ref{tab:expert-skills} lists every Expert Skill using the name and description declared in its \texttt{SKILL.md} header.

\begin{orbenchtable}
\begin{longtable}{@{}p{0.34\linewidth}p{0.61\linewidth}@{}}
\caption{Expert skills available in the skill-enhancement experiment.}
\label{tab:expert-skills}\\
\toprule
\textbf{Skill} & \textbf{Description} \\
\midrule
\endfirsthead
\multicolumn{2}{l}{\small\itshape Table~\thetable\ continued from the previous page.}\\
\toprule
\textbf{Skill} & \textbf{Description} \\
\midrule
\endhead
\midrule
\multicolumn{2}{r}{\small\itshape Continued on the next page.}\\
\endfoot
\bottomrule
\endlastfoot
\nolinkurl{column-generation} & Formulate, implement, stabilize, and debug Dantzig--Wolfe decomposition, restricted master problems, pricing subproblems, column generation, branch-and-price, and integer restricted-master heuristics for cutting stock, set covering, set partitioning, routing, crew scheduling, packing, production planning, and other models with too many variables to enumerate. Use when columns represent complete patterns, routes, schedules, duties, or plans and missing variables can be priced from master dual values. \\
\orstripe
\nolinkurl{constraint-programming} & Model and solve discrete-domain feasibility, scheduling, sequencing, allocation, rostering, and logical problems using constraint-programming concepts, global constraints, propagation, search, and executable PySCIPOpt fallback encodings. Use for all-different, no-overlap, cumulative resource, optional interval, table, element, circuit, reservoir, precedence, calendar, and reified constraints, especially when logical scheduling structure dominates continuous linear economics. \\
\nolinkurl{metaheuristic-optimization} & Design, implement, tune, and validate constructive heuristics, local search, simulated annealing, tabu search, genetic algorithms, variable-neighborhood search, ant-colony methods, particle-swarm methods, and hybrid matheuristics for large optimization problems. Use when exact MIP is too slow, a high-quality feasible incumbent is needed quickly, the task explicitly requests a metaheuristic, or domain-specific operators can exploit routing, scheduling, packing, assignment, location, or continuous decision structure. \\
\orstripe
\nolinkurl{multi-objective-optimization} & Formulate, solve, and compare optimization problems with conflicting objectives using payoff tables, normalized weighted sums, epsilon-constraint sweeps, lexicographic optimization, goal programming, achievement scalarization, Pareto dominance, nondominated sorting, and preference-based selection. Use when a task balances cost, service, emissions, resilience, fairness, risk, quality, or other competing metrics and requires defensible trade-off solutions rather than one arbitrary blended objective. \\
\nolinkurl{network-flow-optimization} & Formulate, solve, decompose, export, and independently validate network flow and network design models with PySCIPOpt or exact graph algorithms available in the runtime. Use for shortest path, maximum flow, minimum-cost flow, transportation, transshipment, circulation, lower-bounded flow, multi-source or multi-sink flow, multi-commodity routing, shared arc capacity, fixed-charge network design, bandwidth allocation, empty repositioning, evacuation, energy or water networks, and time-expanded networks. \\
\orstripe
\nolinkurl{or-diagnosis-repair} & Diagnose and repair failed or unreliable operations-research implementations using evidence rather than symptom lists. Use when optimization code crashes, data and tuple keys disagree, a solver reports infeasible or unbounded, no incumbent is found, solution quality or runtime is poor, numerical behavior is suspicious, or exported decisions fail a checker or downstream system. Supports solver-agnostic diagnosis with reusable PySCIPOpt instrumentation for collecting model, search, IIS, solution, and artifact evidence. \\
\nolinkurl{or-workflow} & Execute operations-research tasks end to end, from decision and data understanding through formulation, implementation, exploratory solves, diagnosis, improvement, and delivery. Use when a task requires a complete optimization workflow rather than an isolated model, solver snippet, or debugging answer. Adapt the rigor to the decision risk, data size, solver environment, time budget, and expected deliverable. \\
\orstripe
\nolinkurl{routing-optimization} & Formulate, solve, improve, export, and independently validate vehicle-routing and dispatch models with PySCIPOpt. Use for TSP, CVRP, VRPTW, pickup-and-delivery, dial-a-ride, multi-depot and heterogeneous fleets, split deliveries, backhauls, compartments, charging or refueling, route-dependent state, rolling replanning, home-healthcare routing, service technicians, evacuation, maritime or multimodal routing, and other tasks where ordered visits must form feasible routes. \\
\nolinkurl{stochastic-optimization} & Formulate, implement, and validate optimization under uncertainty using scenario-based extensive forms, two-stage and multistage stochastic programs, nonanticipativity, chance constraints, sample-average approximation, expected value, CVaR, robust counterparts, scenario trees, and decomposition. Use when demand, supply, travel time, price, capacity, yield, lead time, or disruptions are uncertain and decisions occur before and after information is revealed. \\
\end{longtable}
\end{orbenchtable}

\subsubsection{Observed Expert-Skill Utilization}

We audit the Codex trajectories using a strict invocation criterion: a skill is counted only when a successful command reads its \texttt{SKILL.md}; mentions in reasoning or failed file reads do not count.
Across 107 tasks, 106 tasks successfully read at least one Expert Skill, producing 151 successful read events.
Nine initial reads use an incorrect system-skill path; eight tasks recover through the installed Expert-Skill path, while one task never completes a successful skill read.
The general \texttt{or-workflow} skill dominates usage, covering 105 tasks and 114 reads.
Only 35 tasks invoke a structure-specific skill: 6 Easy, 12 Medium, and 17 Hard tasks.
Thus, specialized-skill routing becomes more frequent with difficulty, rising from 18.75\% of Easy tasks to 50.00\% of Hard tasks.

\begin{table}[H]
\centering
\caption{Observed Expert-Skill utilization in the Codex + GPT-5.4 Expert Skills run. A task is counted once per skill, while reads count repeated successful \texttt{SKILL.md} accesses across agent sessions.}
\label{tab:expert-skill-usage}
\begin{orbenchcompact}
\small
\setlength{\tabcolsep}{5pt}
\begin{tabular}{lrrrrr}
\toprule
\textbf{Skill} & \textbf{Easy} & \textbf{Medium} & \textbf{Hard} & \textbf{Tasks} & \textbf{Reads} \\
\midrule
\texttt{or-workflow} & 32 & 39 & 34 & 105 & 114 \\
\orstripe
\texttt{stochastic-optimization} & 2 & 3 & 10 & 15 & 15 \\
\texttt{routing-optimization} & 0 & 6 & 6 & 12 & 12 \\
\orstripe
\texttt{network-flow-optimization} & 3 & 1 & 1 & 5 & 5 \\
\texttt{or-diagnosis-repair} & 1 & 1 & 1 & 3 & 3 \\
\orstripe
\texttt{column-generation} & 0 & 1 & 0 & 1 & 1 \\
\texttt{metaheuristic-optimization} & 0 & 0 & 1 & 1 & 1 \\
\orstripe
\texttt{constraint-programming} & 0 & 0 & 0 & 0 & 0 \\
\texttt{multi-objective-optimization} & 0 & 0 & 0 & 0 & 0 \\
\bottomrule
\end{tabular}
\end{orbenchcompact}
\end{table}

Usage is highly concentrated.
Seventy-one tasks use only \texttt{or-workflow}; 33 use the workflow plus one specialized skill; one uses four skills; one invokes only \texttt{routing-optimization}; and one has no successful read after an incorrect skill path.
Among the 37 specialized-skill reads, stochastic optimization, routing, and network flow account for 32 (86.49\%).
No trajectory invokes constraint programming or multi-objective optimization.
This indicates that availability alone does not ensure broad method selection: the agent relies heavily on a general workflow and routes to a narrow subset of specialized methods. These figures diagnose skill selection and execution behavior; they do not identify the causal effect of an individual skill.

\subsubsection{Behavioral Mechanisms}
\label{app:expert-skill-mechanisms}

We compare paired trajectories and retained submission artifacts to identify how Expert Skills alter behavior.
Command executions, solve attempts, file-changing commands, and elapsed time are descriptive process measures extracted from the Codex event stream; they do not measure reasoning quality directly.
Across all 107 tasks, Expert Skills add 3.37 command executions and 0.90 minutes per task on average, but reduce file-changing commands by 0.22.
The aggregate effect hides a sharp split.
For the eight pass gains, the paired increases are 12.88 commands, 1.38 solve attempts, and 3.05 minutes.
For the twelve pass losses, they are only 1.75 commands, 0.67 solve attempts, and 0.03 minutes, with 1.58 fewer file-changing commands.
Thus, gains occur when guidance expands diagnosis and solution search; losses occur when it changes the framing without producing enough corrective iteration.

\begin{table}[H]
\centering
\caption{Representative mechanisms from paired trajectories and retained artifacts. Quality values are normalized to $[0,1]$. Internal validation refers to checks implemented by the agent, not the hidden evaluator.}
\label{tab:expert-skill-mechanisms}
\begin{orbenchcompact}
\scriptsize
\setlength{\tabcolsep}{3pt}
\begin{tabularx}{\linewidth}{@{}lccX@{}}
\toprule
\textbf{Task} & \textbf{Without} & \textbf{Expert} & \textbf{Observed mechanism} \\
\midrule
\texttt{IndustryOR\_96}
& $q=.074$
& $q=.746$
& Pattern guidance leads to dominance filtering and a 184-pattern restricted exact solve, which reaches optimality in 8.4\,s. The baseline uses 912 pattern--shift options and terminates with a 13.2\% gap. \\
\orstripe
\texttt{real\_world\_3d\_bin\_packing}
& No solution
& $q=.612$
& Both runs seek structured stack templates, but the baseline leaves no solution artifact. The Expert run completes a restricted template, writes all 38 placements, and checks geometry, support, stack load, and unloading before export. \\
\texttt{railway\_disruption\_recovery}
& Infeasible
& $q=.915$
& The baseline's local checker accepts the export, but the hidden evaluator finds dwell violations on all 14 trains. The Expert run preserves the integrated model, spends the full solve budget, and exports event times consistent with the task convention. \\
\orstripe
\texttt{IndustryOR\_14}
& $q=1.000$
& Infeasible
& The Expert run reports 516 passing self-checks and an objective above the verified upper bound, yet violates the Tuesday chicken production cap. The local validator reproduces the model's omission instead of independently testing the business rule. \\
\texttt{cycling\_network\_design}
& $q=1.000$
& $q=0$
& The baseline enumerates scenario paths and solves for 257\,s to optimality. The network-flow-guided run limits its SCIP phase to about 60\,s, falls back to a heuristic, and loses 18.6\% of the accessibility objective. \\
\orstripe
\texttt{electric\_medical\_waste\_lrp}
& $q=1.000$
& Infeasible
& The Expert artifact declares all facility-window checks passed, but the hidden evaluator finds kilogram-cap violations at two midday windows. The failure is semantic validator mismatch, not solver infeasibility. \\
\texttt{stochastic\_surgery\_capacity}
& $q=.843$
& $q=.357$
& The stochastic-guided model opens one fewer block and lowers first-stage cost, but exact evaluation shows substantially higher expected and tail-risk costs. Optimizing the tractable proxy does not preserve ranking under the benchmark objective. \\
\bottomrule
\end{tabularx}
\end{orbenchcompact}
\end{table}

Three mechanisms follow.
First, skills are effective as \emph{search controllers}: they improve outcomes when they cause the agent to restrict a weak formulation, retain a feasible incumbent, or spend additional iterations on the binding structure.
Second, they are unreliable as \emph{semantic guarantees}: an agent-authored validator often shares the same interpretation error as the model, so a detailed ``PASS'' log can amplify false confidence.
Third, specialized skills can create \emph{objective and budget displacement}: more effort goes into a canonical formulation or surrogate, leaving less time for exact-score improvement and artifact-level repair.
The design implication is direct: future skills should require independent constraint tests, compare surrogate and realized objectives, and trigger repair when a claimed optimum conflicts with an external bound or operational invariant.

\section{Prompt Templates}
\label{app:prompt}

The benchmark uses separate single-step and multi-step instruction regimes. All 99 single-step tasks share the following complete instruction, apart from one inconsequential comma variant. We reproduce it rather than normalize it because the solver settings, iteration budget, and method-selection guidance are part of the evaluated contract.

\begin{orappendixbox}{Single-step agent instruction (common template)}
\begin{lstlisting}[style=orbenchcode,basicstyle=\ttfamily\scriptsize]
You are solving an operations-research benchmark task inside a sandbox.

Work only inside `/app`. Do not use the internet, hidden evaluator files, reference models, package registries, or remote data.

Required workflow:
1. Read `/app/PROBLEM_STATEMENT.md` and every file under `/app/data/`.
2. Write a complete mathematical model to `/app/submissions/model.md` before finalizing code.
3. Implement the solver in `/app/submissions/solve.py` using PySCIPOpt.
4. Use SCIP relative gap `0.0005`. If `ORCLAW_SOLVE_TIME_LIMIT_SECONDS` is set, use a time limit no larger than that value. Preserve solution values with precision up to `1e-8` after the decimal point.
5. Make `/app/submissions/solve.py` solve from scratch when run as `python /app/submissions/solve.py`.
6. Write the required solution file under `/app/submissions/` using the task schema.
7. Write `/app/submissions/solve_log.md` with commands, solver status, objective or score if available, and validation checks.

Note:
You may improve the modeling and solving strategy based on preliminary results, but each individual solve attempt is limited to 5 minutes. The whole workflow is also time-limited, so do not perform unlimited full solve-and-iterate cycles.

Before running expensive solves, reason about the problem structure and choose an efficient approach. You may:
1. Use stronger mathematical formulations and more sophisticated modeling strategies tailored to the structure, special properties, and scale of the given problem instance.
2. Solve directly with SCIP when the model is tractable.
3. Design effective heuristic, local-search, repair, or rounding methods when exact optimization is too slow.
4. Combine heuristics with SCIP, for example by generating an initial feasible solution, fixing or relaxing selected variables, using warm starts if supported, or solving restricted subproblems.
5. Run short diagnostic solves to estimate difficulty, then refine the model or solution method.

Do not default to a naive formulation. Before coding, explicitly analyze whether the problem is better represented as an assignment model, network flow model, set partitioning model, time-indexed model, interval/order-based scheduling model, routing model, or decomposition-based model and so on. Choose the formulation that is likely to give the strongest relaxation and fastest solution for the given instance size. Use preprocessing, dominance filtering, bound tightening, valid inequalities, symmetry breaking, and heuristic warm starts whenever applicable.

Use the limited iteration budget wisely:
- First ensure feasibility and correct output schema.
- Then improve objective quality.
- Prefer targeted refinements over repeated full solves.
- Stop iterating when additional improvements are unlikely within the remaining time.
- Record every meaningful solve attempt and validation result in `/app/submissions/solve_log.md`.

The final submission should prioritize correctness, feasibility, and robust execution from scratch under the time limit.
\end{lstlisting}
\end{orappendixbox}

\paragraph{Multi-step instruction template.}
The 25 multi-step instructions share a shorter event-conditioned contract. The template below retains their common content; double braces denote task- or step-specific substitutions, and optional state or baseline clauses are included only when the step requires them.

\begin{orappendixbox}{Multi-step agent instruction (generalized template)}
\begin{lstlisting}[style=orbenchcode,basicstyle=\ttfamily\scriptsize]
You are {{CURRENT_STAGE_OR_TIME}}. {{EVENT_OR_DECISION_CONTEXT}}

Work only inside `/app`. Read `/app/PROBLEM_STATEMENT.md`, every file under `/app/data/`, and {{OPTIONAL_EVENT_NOTICE_OR_STATE_FILES}}.

Required outputs:

- `/app/submissions/{{MODEL_FILE}}`
- `/app/submissions/{{SOLVER_FILE}}`
- `/app/submissions/{{DECISION_FILE}}`
- `/app/submissions/{{SOLVE_LOG_FILE}}`

Use {{SOLVER_OR_METHOD_REQUIREMENT}} and the {{OUTPUT_FORMAT}} schema described in `/app/PROBLEM_STATEMENT.md`.

{{INFORMATION_BOUNDARY}}
{{REPLANNING_HORIZON_AND_BASELINE}}
{{FROZEN_COMMITMENTS_AND_CARRIED_STATE}}
{{NEW_EVENT_SPECIFIC_REQUIREMENTS}}

This step is time-limited. First produce a correct feasible {{PLAN_OR_REPLAN}}, then improve {{OBJECTIVE}} with {{STRUCTURE_APPROPRIATE_METHODS}} if time remains.

{{SCIP_GAP_TIME_LIMIT_AND_PRECISION_NOTE}}
\end{lstlisting}
\end{orappendixbox}

For an initial step, the information boundary states that future events are not visible. For a later step, the replanning clauses identify the previous accepted submission, persisted state, remaining horizon, frozen decisions, and committed pipeline actions. If an event changes the operational model, the instruction also requires the agent to add the new decision variables or policies rather than merely rerun the previous solver.

\section{Task Examples and Case Studies}
\label{app:case-studies}
\raggedbottom

The following examples connect the benchmark packet, the six construction-time difficulty dimensions, and observed agent behavior. Case trajectories are selected only from retained jobs with complete verifier evidence; they illustrate mechanisms rather than estimate model-level performance. All reported feasibility and quality values are produced by the hidden evaluator. The constraint checklists highlight representative major constraint families rather than every verifier check.

\subsection{Easy Task: Additive Microfactory Order Planning}

\begin{oroverviewbox}{Task overview: \texttt{additive\_microfactory\_order\_planning} (Easy, L2)}
\begin{minipage}[t]{0.57\linewidth}
\vspace{0pt}
\textbf{Operational setting.}
A distributed additive-manufacturing microfactory allocates near-term customer
orders across printers. Printers differ in technology, qualified materials,
speed, available hours, setup burden, operating cost, and medical
certification.

\smallskip
\textbf{Decision and objective.}
The agent activates printer--material setups and assigns integer part quantities
to eligible printers. It maximizes contribution profit after printing,
scrap-adjusted material, and setup costs, while every order must meet its
published minimum fill rate.

\smallskip
\textbf{Instance and difficulty.}
The current instance has 5 printers, 5 materials, 20 orders, and 15 eligible
printer--material pairs. It yields a compact fixed-charge production MIP with
setup binaries and integer production variables. The optimization is small;
the main challenge is preserving the quantifier in the public service rule:
the minimum applies to every order, not only to orders the model elects to
accept.
\end{minipage}\hfill
\begin{minipage}[t]{0.39\linewidth}
\vspace{0pt}
\centering
\begin{tikzpicture}[
  node distance=0.32cm,
  item/.style={draw=ORCaseInfo,fill=white,align=center,text width=0.82\linewidth,
    minimum height=0.72cm,font=\scriptsize},
  arrow/.style={-{Latex[length=1.6mm]},draw=ORCaseInfo}
]
\node[item] (a) {Orders, printers, materials,\\eligibility and costs};
\node[item,below=of a] (b) {Setup binaries + integer\\printer allocations};
\node[item,below=of b] (c) {Highest-profit feasible\\production CSV};
\draw[arrow] (a)--(b);
\draw[arrow] (b)--(c);
\end{tikzpicture}

\vspace{3pt}
\begin{tabular}{@{}lc@{}}
\toprule
\textbf{Difficulty dimension} & \textbf{Level}\\
\midrule
Solving strategy & \ordimbar{0}\\
Formulation & \ordimbar{1}\\
Constraint coupling & \ordimbar{1}\\
Dynamic structure & \ordimbar{0}\\
Data scale & \ordimbar{1}\\
Problem understanding & \ordimbar{1}\\
\bottomrule
\end{tabular}
\end{minipage}
\end{oroverviewbox}

\begin{ororaclebox}
\orcheckcell{1}{Mandatory order fill}{
For every order, total production across eligible printers must be at least its
minimum-fill fraction of requested parts and no greater than the request.}
\hfill
\orcheckcell{2}{Eligibility and certification}{
Production may use only qualified printer--material pairs, and regulated
orders may use only medically certified printers.}

\medskip
\orcheckcell{3}{Setup and printer capacity}{
Every positive production assignment needs its printer--material setup.
Setup time plus processing time must stay within each printer's hours.}
\hfill
\orcheckcell{4}{Material and integrality}{
Scrap-adjusted material use must respect qualified inventory; setup decisions
are binary and produced part counts are integral.}
\end{ororaclebox}

\subsubsection*{Representative trajectories}
\begin{minipage}[t]{0.49\linewidth}
\vspace{0pt}
\begin{ortrajectorybox}{GPT-5.4: failure ($F=0$, $q=0$)}
\begin{enumerate}[leftmargin=*,nosep,label=\arabic*.]
\item Correctly modeled setup activation, printer hours, scrap-adjusted material use, medical certification, and integer production.
\item Added an order-acceptance binary $y_o$ and replaced the public lower bound
with $\sum_p x_{op}\ge L_o y_o$, allowing an order to be rejected with zero
production.
\item Its independent checker repeated the same interpretation, accepting each
order quantity if it was either zero or within its fill interval.
\item The resulting plan omitted O02, O07, and O18, reported profit
12,305.9483, and passed all of its local checks.
\end{enumerate}
\textbf{Verifier evidence:} three hard errors,
\texttt{O02/O07/O18 misses required fill rate}; the plan was infeasible, so
quality was zero.
\end{ortrajectorybox}
\end{minipage}\hfill
\begin{minipage}[t]{0.49\linewidth}
\vspace{0pt}
\begin{ortrajectorybox}{DeepSeek-V4-Pro: success ($F=1$, $q=2$)}
\begin{enumerate}[leftmargin=*,nosep,label=\arabic*.]
\item Kept the fill constraint unconditional for every order:
$\lceil \rho_o D_o\rceil\le\sum_p x_{op}\le D_o$.
\item Linked every production variable to a qualified setup, filtered regulated
orders to certified printers, and modeled printer-hour and material capacities.
\item Solved the compact MIP to the requested relative gap and activated seven
printer--material setups.
\item Submitted positive production for all 20 orders; the hidden evaluator
verified 628 parts, zero errors, and profit 11,688.0909.
\end{enumerate}
\textbf{Verifier evidence:} zero errors, feasibility one, and maximum quality
score $q=2$.
\end{ortrajectorybox}
\end{minipage}

\begin{minipage}[t]{0.49\linewidth}
\vspace{0pt}
\begin{orfailurebox}{Failure analysis}
The public packet states that \emph{every order} has a minimum fill rate and
that shipped quantity for that order must reach the stated fraction. The failed
trajectory changed this mandatory service requirement into an optional
accept-or-reject decision. Its validator then checked the modified model rather
than the public contract, so solver status and local feasibility agreed with
each other while both disagreed with the task.
\end{orfailurebox}
\end{minipage}\hfill
\begin{minipage}[t]{0.49\linewidth}
\vspace{0pt}
\begin{orsuccessbox}{Success analysis}
The successful trajectory preserved the universal quantifier and used the
minimum-fill values directly as order-level lower bounds. It then optimized
allocation and setup choices inside the remaining feasible region. The
contrast is not computational: both agents solved small MIPs. It is semantic.
A locally coherent formulation can still fail when it silently introduces a
business decision that the public specification does not permit.
\end{orsuccessbox}
\end{minipage}

\clearpage
\subsection{Medium Task: Scenario-Aware Cycling Network Design}

\begin{oroverviewbox}{Task overview: \texttt{cycling\_network\_design} (Medium, L4)}
\begin{minipage}[t]{0.57\linewidth}
\vspace{0pt}
\textbf{Operational setting.}
A city selects cycling-link upgrades and intersection treatments to create a usable low-stress network. Residents use shortest feasible routes, and accessibility decreases piecewise with travel time. Four resilience scenarios remove corridors or facility types before routes are recomputed.

\smallskip
\textbf{Decision and objective.}
The binary portfolio must maximize scenario-weighted access to jobs and services while satisfying separate edge and node budgets, total capital and project-count limits, equity share, dependencies, bridge-feeder rules, downtown disruption limits, and corridor commitments.

\smallskip
\textbf{Instance and difficulty.}
The instance has 30 nodes, 106 directed edges, 45 projects, and 348 OD pairs. A project changes edge and node passability, which changes shortest paths, which in turn changes the portfolio objective in every scenario. This endogenous network evaluation explains the high formulation, coupling, and solving-strategy scores; the task is not well served by a static project-value approximation.
\end{minipage}\hfill
\begin{minipage}[t]{0.39\linewidth}
\vspace{0pt}
\centering
\begin{tikzpicture}[
  node distance=0.28cm,
  item/.style={draw=ORCaseInfo,fill=white,align=center,text width=0.84\linewidth,
    minimum height=0.67cm,font=\scriptsize},
  arrow/.style={-{Latex[length=1.6mm]},draw=ORCaseInfo}
]
\node[item] (a) {Select infrastructure\\portfolio};
\node[item,below=of a] (b) {Recompute usable network\\under each scenario};
\node[item,below=of b] (c) {Shortest paths $\rightarrow$\\weighted accessibility};
\draw[arrow] (a)--(b);
\draw[arrow] (b)--(c);
\end{tikzpicture}

\vspace{3pt}
\begin{tabular}{@{}lc@{}}
\toprule
\textbf{Difficulty dimension} & \textbf{Level}\\
\midrule
Solving strategy & \ordimbar{3}\\
Formulation & \ordimbar{3}\\
Constraint coupling & \ordimbar{3}\\
Dynamic structure & \ordimbar{2}\\
Data scale & \ordimbar{2}\\
Problem understanding & \ordimbar{2}\\
\bottomrule
\end{tabular}
\end{minipage}
\end{oroverviewbox}

\begin{ororaclebox}
\orcheckcell{1}{Resource budgets}{
Selected edge upgrades must respect the edge-kilometre-equivalent budget;
node treatments have a separate budget; and total capital expenditure must
remain within the citywide cap.}
\hfill
\orcheckcell{2}{Portfolio composition}{
The portfolio cannot exceed the total project-count limit or the cap on
downtown-disruption projects. Equity-priority projects must receive at least
the required fraction of selected capital expenditure.}

\medskip
\orcheckcell{3}{Dependencies and bridge feeders}{
Whenever a selected project declares a prerequisite, that prerequisite must
also be selected. Each selected bridge project must be accompanied by the
minimum number of eligible feeder projects on every specified side.}
\hfill
\orcheckcell{4}{Corridor commitments}{
For every corridor commitment, the number of selected projects belonging to
that corridor must reach its prescribed minimum. A portfolio violating any
budget, composition, dependency, feeder, or corridor rule is infeasible.}
\end{ororaclebox}

\subsubsection*{Representative trajectories}
\begin{minipage}[t]{0.49\linewidth}
\vspace{0pt}
\begin{ortrajectorybox}{Qwen3.6 Plus: feasible, below threshold ($F=1$, $q=0$)}
\begin{enumerate}[leftmargin=*,nosep,label=\arabic*.]
\item Built feasible 27-project portfolios with MILP seeds and greedy project swaps.
\item Coded impedance as flat through 20 minutes, then used only the shallow first slope through 40 minutes.
\item Used that routine to rank every move and validate the final portfolio.
\item Reported 300,734,176 after side-constraint checks, without independent objective validation.
\end{enumerate}
\textbf{Verifier evidence:} exact objective 204,851,681 versus reference 216,495,022; replaying Qwen's formula on the verifier paths exactly reproduces its reported 300,734,176.
\end{ortrajectorybox}
\end{minipage}\hfill
\begin{minipage}[t]{0.49\linewidth}
\vspace{0pt}
\begin{ortrajectorybox}{Kimi K2.6: success ($F=1$, $q=1$)}
\begin{enumerate}[leftmargin=*,nosep,label=\arabic*.]
\item Rejected a full path formulation as unnecessarily large and initialized a restricted path set.
\item Solved a project-selection master problem using those candidate routes.
\item Evaluated the incumbent with exact shortest paths, detected missing improving routes, and added them.
\item Re-optimized to convergence and polished the portfolio with exact-objective one-swap checks.
\end{enumerate}
\textbf{Verifier evidence:} feasible portfolio and objective 216,495,021.89, matching the verified reference.
\end{ortrajectorybox}
\end{minipage}

\begin{minipage}[t]{0.49\linewidth}
\vspace{0pt}
\begin{orfailurebox}{Failure analysis}
This was an objective-function specification error, not a feasibility or weak-search failure. Qwen treated the first breakpoint as a zero-penalty plateau and never used the steeper second slope. The correct function declines from time zero and steepens after 20 minutes; at 30 minutes, Qwen assigned 0.85 instead of 0.35. Its search therefore ignored improvements among sub-20-minute trips and greatly overvalued medium-length routes. Because the same faulty scoring oracle ranked moves and checked the final result, repeated optimization systematically improved the wrong objective while retaining a feasible but subthreshold portfolio.
\end{orfailurebox}
\end{minipage}\hfill
\begin{minipage}[t]{0.49\linewidth}
\vspace{0pt}
\begin{orsuccessbox}{Success analysis}
Kimi separated portfolio optimization from exact network evaluation and placed them in a refinement loop. Restricted paths kept the master problem tractable, while repeated shortest-path recomputation exposed objective terms omitted by the current model. The decisive behavior was not a particular solver choice, but evaluator-aligned search: every claimed improvement was tested under the same network semantics used for final grading.
\end{orsuccessbox}
\end{minipage}

\clearpage
\subsection{Hard Task: Stochastic Surgery Capacity Planning}

\begin{oroverviewbox}{Task overview: \texttt{stochastic\_surgery\_capacity\_planning} (Hard, L5)}
\begin{minipage}[t]{0.57\linewidth}
\vspace{0pt}
\textbf{Operational setting.}
A hospital plans the next week's elective surgeries while emergency arrivals and procedure durations remain uncertain. The schedule must remain clinically and operationally valid across operating rooms, surgeons, anesthesia teams, specialty equipment, PACU, wards, and ICU capacity.

\smallskip
\textbf{Decision and objective.}
For each of 36 cases, the agent schedules or defers the case, assigns one of 12 compatible blocks, and chooses a sequence and start time. The objective combines fixed opening, deferment, and lateness costs with expected scenario overtime, idle time, recovery overflow, and CVaR tail risk over 12 scenarios.

\smallskip
\textbf{Instance and difficulty.}
Assignment decisions determine temporal overlap, shared-resource use, and downstream occupancy. Scenario durations then propagate through the submitted sequence and alter room completion and recovery demand. This creates dense coupling between first-stage scheduling and risk-aware recourse; a feasible fallback must still satisfy exact output and sequencing contracts.
\end{minipage}\hfill
\begin{minipage}[t]{0.39\linewidth}
\vspace{0pt}
\centering
\begin{tikzpicture}[
  node distance=0.28cm,
  item/.style={draw=ORCaseInfo,fill=white,align=center,text width=0.84\linewidth,
    minimum height=0.67cm,font=\scriptsize},
  arrow/.style={-{Latex[length=1.6mm]},draw=ORCaseInfo}
]
\node[item] (a) {Assign, sequence, and\\time elective cases};
\node[item,below=of a] (b) {Propagate 12 duration and\\emergency scenarios};
\node[item,below=of b] (c) {Expected operating cost\\+ tail-risk penalty};
\draw[arrow] (a)--(b);
\draw[arrow] (b)--(c);
\end{tikzpicture}

\vspace{3pt}
\begin{tabular}{@{}lc@{}}
\toprule
\textbf{Difficulty dimension} & \textbf{Level}\\
\midrule
Solving strategy & \ordimbar{2}\\
Formulation & \ordimbar{3}\\
Constraint coupling & \ordimbar{3}\\
Dynamic structure & \ordimbar{3}\\
Data scale & \ordimbar{2}\\
Problem understanding & \ordimbar{3}\\
\bottomrule
\end{tabular}
\end{minipage}
\end{oroverviewbox}

\begin{ororaclebox}
\orcheckcell{1}{Case--block eligibility}{
Every case is either scheduled once or deferred. A scheduled case must use a
block after its ready day with the required specialty, pediatric, cardiac, and
ICU-direct capabilities.}
\hfill
\orcheckcell{2}{Block sequence and timing}{
Sequences must be unique within a block. Case count and high-acuity limits
must hold; at most one infection-control case is allowed and it must be last.
Starts must respect setup, turnover, and the planned overtime buffer.}

\medskip
\orcheckcell{3}{Anesthesia and surgeons}{
Daily anesthesia-type limits must hold. Each primary surgeon must be
credentialed and available, remain within daily case and minute limits, and
have no overlapping planned intervals across different rooms.}
\hfill
\orcheckcell{4}{Shared equipment}{
Each case must be eligible for its equipment pool. Daily uses cannot exceed
the pool limit, and interval occupancy across all rooms, including equipment
turnaround, cannot exceed the number of available units.}
\end{ororaclebox}

\subsubsection*{Representative trajectories}
\begin{minipage}[t]{0.49\linewidth}
\vspace{0pt}
\begin{ortrajectorybox}{Claude Opus 4.6: failure ($F=0$, $q=0$)}
\begin{enumerate}[leftmargin=*,nosep,label=\arabic*.]
\item Built a deterministic-equivalent assignment model with scenario costs and CVaR, reporting a 0.00\% MIP gap.
\item Enforced equipment use counts by day and added turnaround corrections for same-pool cases within one OR block.
\item Sequenced and validated each block independently, then reported 34 of 36 cases scheduled with zero internal errors.
\item Assigned S023 and S024 to different day-3 rooms, both at minute 15, although both require the single-unit \texttt{robotic\_scope}.
\end{enumerate}
\textbf{Verifier evidence:} one hard-constraint error: \texttt{robotic\_scope} occupancy reached 2 on day 3 although only one unit was available.
\end{ortrajectorybox}
\end{minipage}\hfill
\begin{minipage}[t]{0.49\linewidth}
\vspace{0pt}
\begin{ortrajectorybox}{GPT-5.3-Codex: success ($F=1$, $q=0.921$)}
\begin{enumerate}[leftmargin=*,nosep,label=\arabic*.]
\item Used a compact assignment, interval, and ordering formulation for blocks and shared resources.
\item Integrated expected scenario cost and CVaR while retaining a valid incumbent during improvement.
\item Scheduled 34 of 36 cases and preserved the sequence and start-time semantics required by the evaluator.
\item Re-parsed the final artifact and independently checked schema and hard feasibility.
\end{enumerate}
\textbf{Verifier evidence:} zero errors; objective 110,193.65 versus reference 111,966.26, yielding normalized quality 0.921.
\end{ortrajectorybox}
\end{minipage}

\begin{minipage}[t]{0.49\linewidth}
\vspace{0pt}
\begin{orfailurebox}{Failure analysis}
The submission contract was valid and the evaluator reconstructed the full schedule. The failure instead came from an incomplete shared-resource model: daily equipment-use limits and within-block turnaround corrections do not prevent simultaneous use across different rooms. Consequently, SCIP proved optimality only for a relaxation that omitted cross-block equipment occupancy. The run's independent block-level validation repeated the same omission and certified an infeasible schedule. This case shows that hard scheduling tasks require time-coupled resource constraints across all concurrent rooms, not only aggregate counts or local sequencing checks.
\end{orfailurebox}
\end{minipage}\hfill
\begin{minipage}[t]{0.49\linewidth}
\vspace{0pt}
\begin{orsuccessbox}{Success analysis}
GPT-5.3-Codex combined a formulation compact enough to produce useful incumbents with explicit preservation of feasibility and output semantics. Its final objective improved on the bundled reference while remaining close to the verified global bound. The case illustrates a three-part success pattern for hard OR tasks: choose a tractable representation, maintain a valid fallback throughout search, and validate the exact exported decisions before submission.
\end{orsuccessbox}
\end{minipage}

\clearpage
\subsection{Multi-Step Task: Online Liner Slot and Empty-Container Repositioning}

Under the strict task-level criterion, a multi-step task passes only when the
complete trajectory is feasible; a strong early step cannot compensate for a
later infeasible replan. Five of the eight multi-step tasks were passed by at
least one evaluated agent. We select the following task because both a complete
success and a complete failure trajectory were retained.

\begin{oroverviewbox}{Task overview: \texttt{online\_liner\_slot\_empty\_repositioning} (Hard, L5)}
\textbf{Operational setting.}
An intra-Asia liner carrier receives booking requests over three decision
points. The agent must accept or reject each visible request, route accepted
containers through scheduled sailings, and reposition or lease empty
containers. The objective maximizes revenue net of rejection, repositioning,
leasing, holding, customs, carbon, and plan-change costs.

\smallskip
\textbf{Principal difficulty.}
Loaded paths share vessel TEU, reefer-plug, and dangerous-goods capacity with
empty moves. They also induce time-dependent empty inventories, yard and gate
flows, customs inspections, priority-service floors, and transfer limits.
Medical reefers and protected service blocks add event-specific obligations.

\smallskip
\begin{center}
\begin{tikzpicture}[
  node distance=0.52cm,
  wave/.style={draw=ORCaseInfo,fill=white,rounded corners=1pt,align=center,
    text width=0.26\linewidth,minimum height=1.55cm,font=\small},
  arrow/.style={-{Latex[length=2mm]},thick,draw=ORCaseInfo}
]
\node[wave] (w1) {\textbf{Initial plan, day 0}\\54 bookings\\baseline network plan};
\node[wave,right=of w1] (w2) {\textbf{Replan, day 4}\\88 bookings; freeze $\leq 5$\\labor slowdown and\\reduced leg capacity};
\node[wave,right=of w2] (w3) {\textbf{Final replan, day 7}\\128 bookings; freeze $\leq 8$\\medical-reefer surge and\\Jakarta yard incident};
\draw[arrow] (w1)--(w2);
\draw[arrow] (w2)--(w3);
\end{tikzpicture}
\end{center}

\begin{minipage}[t]{0.58\linewidth}
\vspace{0pt}
\textbf{Why multi-step is different.}
Harbor carries forward the prior submissions and state, then overlays newly
visible bookings, capacities, and event-control tables. The second replan adds
protected slots and temporary reefer/DG1 transfer caps; the final replan
updates these controls and adds direct-sailing rules for medical reefers.
Decisions whose sailings enter the frozen window must retain both acceptance
and vessel path. Thus each artifact must be feasible for the current network
and consistent with the previously approved plan.
\end{minipage}\hfill
\begin{minipage}[t]{0.38\linewidth}
\vspace{0pt}
\centering
\begin{tabular}{@{}lc@{}}
\toprule
\textbf{Difficulty dimension} & \textbf{Level}\\
\midrule
Solving strategy & \ordimbar{3}\\
Formulation & \ordimbar{3}\\
Constraint coupling & \ordimbar{3}\\
Dynamic structure & \ordimbar{3}\\
Data scale & \ordimbar{2}\\
Problem understanding & \ordimbar{3}\\
\bottomrule
\end{tabular}
\end{minipage}
\end{oroverviewbox}

\begin{ororaclebox}
\orcheckcell{1}{Booking service and paths}{
Every visible booking must be accepted or rejected. Accepted paths must be
continuous, depart and arrive within the booking window, obey transfer and
allowed-service rules, and satisfy direct-sailing or mandatory-acceptance
requirements where applicable.}
\hfill
\orcheckcell{2}{Persistent commitments}{
A prior decision and full vessel path are immutable when the first departure
is on or before the inclusive freeze date. An accepted booking that has entered
execution cannot be rejected, and each priority group must meet its minimum
acceptance rate.}

\medskip
\orcheckcell{3}{Vessel and handling capacity}{
Loaded bookings and empty moves jointly consume vessel TEU, reefer-plug, and
DG1 capacity. Protected-slot minima, reefer and DG1 transfer limits, customs
windows, and medical-reefer preclearance limits must all hold.}
\hfill
\orcheckcell{4}{Empty-container feasibility}{
Loaded departures, repositioning, arrivals, and leases are replayed by day and
port. STD and reefer inventories must remain nonnegative; lease availability,
gate-in/out, yard, and reefer-plug capacities cannot be exceeded.}
\end{ororaclebox}

\subsubsection*{Task-level outcome and stage diagnostics}
Stage values below diagnose where the trajectories diverge; only the final
task-level result determines pass or failure.
\begin{center}
\small
\begin{tabular}{@{}lcll@{}}
\toprule
\textbf{Decision point} & \textbf{Visible} &
\textbf{DeepSeek V4 Flash} & \textbf{GPT-5.4}\\
\midrule
Initial plan & 54 & feasible; profit 4,183.42 & feasible; profit 4,183.42\\
Step-2 replan & 88 & feasible; profit 22,321.22 & feasible; profit 19,187.42\\
Final replan & 128 & infeasible; five violations & feasible; profit 43,116.05\\
\midrule
\textbf{Complete trajectory} & -- &
\textbf{Fail} ($F=0$) & \textbf{Pass} ($F=1$, $q=0.807$)\\
\bottomrule
\end{tabular}
\end{center}

\subsubsection*{Representative trajectories}
\begin{minipage}[t]{0.49\linewidth}
\vspace{0pt}
\begin{ortrajectorybox}{DeepSeek V4 Flash: final-step failure ($F=0$)}
\begin{enumerate}[leftmargin=*,nosep,label=\arabic*.]
\item Produced feasible initial and step-2 plans, including path, empty-move,
and inventory decisions for all visible bookings.
\item At the final event, expanded the model to 128 bookings and incorporated
the new medical-reefer and yard-disruption tables.
\item Stated that departures on day 8 were frozen, but implemented the test as
\texttt{depart\_day < 8}. It therefore classified B053 and B061, both routed
on L16 departing on day 8, as changeable and rejected them.
\item Mis-timed customs events in its MIP: it ignored origin preclearance
shifts, charged transfers on the outgoing departure day rather than the
incoming arrival day, and delayed reefer imports by one day. Its internal
checks reused these definitions and therefore accepted an invalid optimum.
\end{enumerate}
\textbf{Verifier evidence:} the five messages represent three physical
violations. B053 and B061 each triggered both a frozen-plan change and an
already-in-execution rejection; the remaining violation was 21 TEU of
cold-chain inspection demand at Singapore on day 10 against a 20-TEU limit.
\end{ortrajectorybox}
\end{minipage}\hfill
\begin{minipage}[t]{0.49\linewidth}
\vspace{0pt}
\begin{ortrajectorybox}{GPT-5.4: complete pass ($F=1$, $q=0.807$)}
\begin{enumerate}[leftmargin=*,nosep,label=\arabic*.]
\item Built a compact path-selection model coupled to empty inventory,
repositioning, leasing, service, and event-specific capacity constraints.
\item Reloaded each prior submission as binding state and explicitly preserved
accepted paths whose departures fell inside the freeze window.
\item Rebuilt candidate paths and operating constraints after each event,
including protected slots, transfer caps, customs windows, and medical rules.
\item Re-parsed the final artifact and compared it with the previous plan
before submission; no frozen acceptance decision or path changed.
\end{enumerate}
\textbf{Verifier evidence:} all three stages had zero errors. The final plan
accepted 121 of 128 bookings, used two empty moves and four leases, incurred no
change cost, and achieved verified profit 43,116.05.
\end{ortrajectorybox}
\end{minipage}

\begin{minipage}[t]{0.49\linewidth}
\vspace{0pt}
\begin{orfailurebox}{Failure analysis}
This is a temporal-semantics formulation error, not weak optimization. The
strict inequality created an off-by-one error at the inclusive freeze boundary:
two binding commitments were released and then rejected. Separately, incorrect
event timestamps moved customs demand away from its prescribed inspection
days, hiding the Singapore overload. The solver optimized the wrong feasible
set, and model-internal validation could not expose the error because it reused
the same faulty time mapping. Multi-step replanning requires exact temporal
indexing of both prior commitments and newly induced resource events.
\end{orfailurebox}
\end{minipage}\hfill
\begin{minipage}[t]{0.49\linewidth}
\vspace{0pt}
\begin{orsuccessbox}{Success analysis}
GPT-5.4 treated persistent artifacts as model input rather than historical
context. It constrained frozen paths before optimizing new bookings, rebuilt
event-dependent capacity accounting, and checked cross-step differences
explicitly. This reduced step-2 profit relative to DeepSeek, but preserved a
feasible base for the final disruption and produced a valid high-quality
complete trajectory.
\end{orsuccessbox}
\end{minipage}

\enlargethispage{3\baselineskip}
\paragraph{Cross-case implication.}
Across these examples, the dominant bottleneck shifts from schema fidelity to evaluator-aligned objectives, formulation and fallback discipline, and finally state continuity. Reliable OR agents must therefore integrate interpretation, optimization, serialization, and independent validation into one workflow.

%%%%%%%%%%%%%%%%%%%%%%%%%%%%%%%%%%%%%%%%%%%%%%%%%%%%%%%%%%%%

\newpage

\end{document}